\documentclass[lettersize,journal]{IEEEtran}
\usepackage{amsmath,amsfonts}
\usepackage{algorithmic}
\usepackage{algorithm}
\usepackage{array}
\usepackage{textcomp}
\usepackage{stfloats}
\usepackage{url}
\usepackage{verbatim}
\usepackage{graphicx}
\usepackage{cite}
\usepackage{color}
\usepackage{hyperref,cleveref}
\usepackage{multirow}
\usepackage{subcaption} 

\usepackage[T1]{fontenc}

\newif\ifappendixincluded
\appendixincludedtrue

\input{mySymbol.sty}

\begin{document}

\title{Attentional Graph Meta-Learning for Indoor Localization Using Extremely Sparse Fingerprints}

\author{Wenzhong~Yan,~\IEEEmembership{Student Member,~IEEE},~Feng~Yin,~\IEEEmembership{Senior Member,~IEEE},\\~Jun~Gao,~\IEEEmembership{Student Member,~IEEE},~Ao~Wang,~Yang~Tian,~Ruizhi~Chen
\thanks{Wenzhong~Yan,~Feng~Yin,~Jun~Gao,~Ao~Wang are with the School of Science \& Engineering, The Chinese University of Hong Kong, Shenzhen, Shenzhen 518172, China (email: wenzhongyan@link.cuhk.edu.cn, yinfeng@cuhk.edu.cn, jungao@link.cuhk.edu.cn, aowang2@link.cuhk.edu.cn).}
\thanks{Yang~Tian is with Huawei Technologies Co., LTD., Shanghai 201206, China (email: yang.tian@huawei.com).}
\thanks{Ruizhi~Chen is with the School of Data Science, The Chinese University of Hong Kong, Shenzhen, Shenzhen 518172, China (email: chenruizhi@cuhk.edu.cn)}
\thanks{(Corresponding Author: Feng Yin, Email: yinfeng@cuhk.edu.cn)}
}

\maketitle

\begin{abstract}

Fingerprint-based indoor localization is often labor-intensive due to the need for dense grids and repeated measurements across time and space. Maintaining high localization accuracy with extremely sparse fingerprints remains a persistent challenge. Existing benchmark methods primarily rely on the measured fingerprints, while neglecting valuable spatial and environmental characteristics. To address this issue, we propose a systematic integration of an Attentional Graph Neural Network (AGNN) model, capable of learning spatial adjacency relationships and aggregating information from neighboring fingerprints, and a meta-learning framework that utilizes datasets with similar environmental characteristics to enhance model training. To minimize the labor required for fingerprint collection, we introduce two novel data augmentation strategies: 1) unlabeled fingerprint augmentation using moving platforms, which enables the semi-supervised AGNN model to incorporate information from unlabeled fingerprints, and 2) synthetic labeled fingerprint augmentation through environmental digital twins, which enhances the meta-learning framework through a practical distribution alignment, which can minimize the feature discrepancy between synthetic and real-world fingerprints effectively. By integrating these novel modules, we propose the Attentional Graph Meta-Learning (AGML) model. This novel model combines the strengths of the AGNN model and the meta-learning framework to address the challenges posed by extremely sparse fingerprints. To validate our approach, we collected multiple datasets from both consumer-grade WiFi devices and professional equipment across diverse environments. These datasets can also serve as a valuable resource for benchmarking fingerprint-based indoor localization methods. Extensive experiments conducted on both synthetic and real-world datasets demonstrate that the AGML model-based localization method consistently outperforms all baseline methods using sparse fingerprints across all evaluated metrics.

\end{abstract}


\section{Introduction}

Location-based services have become crucial in modern society, with various localization techniques extensively studied across multiple scientific domains~\cite{faragher2015location, 6475197,godrich2010target}. While Global Navigation Satellite Systems (GNSS) offer precise localization outdoors, their performance significantly deteriorates in indoor environments due to signal attenuation and interference.
Consequently, fingerprint-based localization has emerged as a promising alternative for indoor environments due to its low hardware cost. 
This approach is typically divided into two main phases: the offline phase and the online phase. During the offline phase, wireless signal attributes, such as Received Signal Strength (RSS) and Channel State Information (CSI), are recorded at designated Reference Points (RPs) to construct a comprehensive fingerprint database. The size of this database is often substantial to ensure adequate coverage. In the online phase, signals captured at a Test Point (TP) are matched to this database to estimate the location, using algorithms like RADAR~\cite{bahl2000radar} and Horus~\cite{youssef2005horus}.

Despite its substantial practical value, fingerprint-based indoor localization faces a critical challenge: the data collection process is often labor-intensive and resource-consuming. Constructing a comprehensive fingerprint database requires extensive site surveys,  during which signal attributes are meticulously measured across multiple locations in any target environment. This requirement is particularly burdensome in large or complex indoor spaces and necessitates ongoing database maintenance as signal conditions fluctuate over time, leading to constantly increasing data collection efforts.

However, existing methods~\cite{bahl2000radar, 7827145, torres2014ujiindoorloc} exhibit notable performance degradation in scenarios with sparse fingerprints, primarily due to their reliance on raw signal measurements without incorporating additional information, such as adjacency relationships between nearby fingerprints or environmental characteristics. The absence of such additional information limits their performance, resulting in suboptimal accuracy and reliability.

Given these limitations, there is an urgent need for an efficient fingerprint-based indoor localization model that can perform well with sparse fingerprints by leveraging additional available information.

\subsection{Related Works}
\label{sec: related_works}

Indoor localization techniques have evolved over many years, and we can classify the existing fingerprinting approaches into classic signal processing-based approaches (both probabilistic and deterministic) and machine learning-based localization.

Probabilistic localization employs statistical analysis to estimate locations by comparing new signal measurements with historical data in a fingerprint database, typically requiring extensive offline calibration. For example, Horus~\cite{youssef2005horus} uses probabilistic models to characterize signal distributions and compute the maximum posterior probability of a target's location. DeepFi~\cite{wang2015deepfi} improves computational efficiency by combining a probabilistic model with a greedy learning algorithm. However, probabilistic methods, while computationally efficient, often struggle in practical environments due to their dependence on precise positional measurements.
Deterministic methods, such as the $K$-Nearest Neighbors (KNN) algorithm, estimate locations based on signal similarity metrics, like Euclidean distance~\cite{wu2017gain} and cosine similarity~\cite{he2014sectjunction}, to identify the nearest fingerprint in the database. Weighted KNN (WKNN) further improves accuracy by assigning weights to neighboring fingerprints based on their proximity to the target location~\cite{WKNN2020indoor}. Despite their simplicity, deterministic methods are vulnerable to signal variability, which may lead to the selection of spatially distant neighbors, thereby reducing localization accuracy.

In recent years, machine learning has revolutionized localization services. Ghozali \textit{et al.} treat indoor localization as a regression problem using RSS data from a real office environment, requiring substantial data for effective training due to random initialization \cite{ghozali2019indoor}. ConFi~\cite{chen2017confi} pioneers the use of Convolutional Neural Networks (CNNs) to learn CSI images at reference points, broadening the scope of indoor localization. Hsieh \textit{et al.} frame localization as a classification problem, leveraging both RSS and CSI data to evaluate various neural network architectures for accurate room-level location estimation  \cite{hsieh2019deep}.

However, the aforementioned machine learning-based localization methods primarily focus on signal measurements from individual fingerprints, neglecting valuable spatial relationships and environmental characteristics. As a result, these methods typically require an extensive fingerprint database for effective model training.

To address this challenge, we explore recent advancements in two key areas: domain-adaptive localization, which leverages datasets with similar environmental characteristics to model training, and Graph Neural Network (GNN)-based localization, which incorporates neighboring information to enhance localization accuracy. Both approaches introduce additional information to improve model training, thereby mitigating the dependence on extensive fingerprint collection. The details are summarized as follows.

\subsubsection{Domain-adaptive Localization}

Domain adaptation techniques, such as transfer learning, are widely employed to address challenges in localization, where the source domain represents the original environment and the target domain is a new, potentially unseen environment with limited data. For instance, TransLoc~\cite{Transloc} utilizes transfer learning to establish cross-domain mappings and create a unified feature space with discriminative information from different domains. Similarly, CRISLoc~\cite{CRISLoc} reconstructs a high-dimensional CSI fingerprint database by combining outdated fingerprints with a few new measurements. Fidora~\cite{chen2022fidora} introduces a domain-adaptive classifier that adjusts to new data using a variational autoencoder and a joint classification-reconstruction framework. ILCL~\cite{zhu2022intelligent} employs incremental learning and expands neural nodes to reduce training time, though it remains prone to overfitting with limited CSI data.

In our previous work~\cite{gao2022metaloc, MetaLoc, 10447939}, we propose MetaLoc, which leverages meta-learning to address environment-dependent limitations in localization. The approach involves training meta-parameters using historical data from diverse, well-calibrated indoor environments. These meta-parameters initialize a deep neural network, enabling rapid adaptation to new environments with minimal data samples.

While the aforementioned methods significantly alleviate data requirements in the new environment, they primarily concentrate on raw signal measurements from fingerprints, neglecting the inherent adjacency information between closely located fingerprints. In the coming era of large-scale networks, the rich information embedded in neighboring fingerprints needs to be further effectively utilized.

\subsubsection{GNN-based Localization}

Graph Neural Networks (GNNs) are a class of neural networks designed to leverage graph structures for extracting meaningful domain knowledge~\cite{xu2018powerful}. Unlike traditional neural networks, which permit unrestricted connections between nodes, GNNs update each node's latent states based on aggregated information from neighboring nodes, enabling the learning of local features within large-scale networks. Localization is an emerging application of GNNs, which forms the core focus of this work. 
In our prior work, we introduced Graph Convolution Networks (GCNs), a GNN variant, for network localization~\cite{yan2021graph}. More recently, we have developed the Attentional GNN (AGNN) model, which incorporates two distinct attention mechanisms into the GCN framework to address sensitivity to thresholding and enhance model capacity~\cite{yan2024attentional}.
Another approach models Access Points (APs) as graph nodes, with edges constructed based on geometric relationships. For example, fully connected graphs are used in~\cite{sun2021novel} and~\cite{kang2023indoor}, with the edge weights inversely proportional to inter-node distances. In~\cite{luo2022geometric}, multiple AP types (e.g., WiFi, ZigBee, Bluetooth) are represented as a fully connected bidirectional graph, with the directional edges between users and APs being weighted inversely to the estimated distances. By leveraging the GraphSAGE layer~\cite{hamilton2017inductive}, users can be localized by unifying generated graphs.

GNN-based localization techniques consistently demonstrate improved accuracy and robustness under varying signal propagation conditions by aggregating information from graph structures. However, they face a practical limitation: the need for a large number of labeled fingerprint samples, which requires labor-intensive and time-consuming data collection. Our prior work~\cite{yan2021graph} addresses this by employing semi-supervised learning, showing that GNNs can enhance localization accuracy by incorporating abundant unlabeled fingerprint samples. This approach offers a promising solution to the challenge of limited labeled data availability.

\begin{figure*}
    \centering
    \includegraphics[width=1\linewidth]{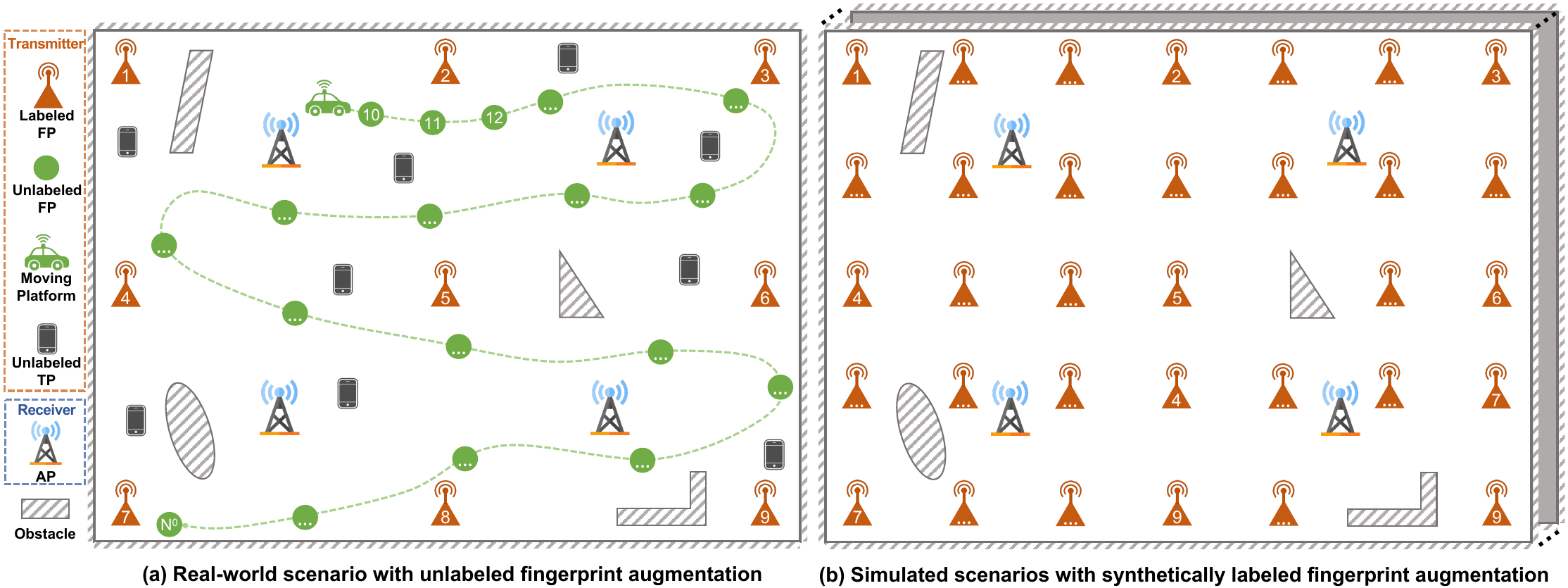}
    \caption{Two fingerprint augmentation strategies based on the original real-world indoor localization scenario.}
    \label{fig: Indoor_Loc_Scenario}
\end{figure*}

\subsection{Contributions}
\label{sec: contributions}

Motivated by these factors, in this paper, we propose an Attentional Graph Meta-Learning (AGML) model that exploits both model adaptability and adjacency relationships to enable robust and accurate localization in real-world scenarios with extremely limited labeled fingerprints. 
The main contributions are summarized as follows.

\begin{itemize}
    \item We extend our previously proposed AGNN model, originally designed for network localization using inter-node distance measurements~\cite{yan2024attentional}, to fingerprint-based indoor localization. To achieve this, we introduce an additional Distance Learning Module (DLM) to effectively process CSI or Channel Impulse Response (CIR) data received from a limited set of APs.

    \item To address the challenge of sparse fingerprints,  we propose a comprehensive data augmentation approach that integrates two strategies: 1) unlabeled fingerprint augmentation using moving platforms to enhance the localization accuracy of the AGNN model, and 2) synthetic labeled fingerprint augmentation through environmental digital twins, coupled with a distribution alignment method to ensure feature consistency between real-world and synthetic data. This unified approach not only enriches fingerprint diversity but also improves the generalization capability of the AGML model.

    \item Our proposed AGML model combines the strengths of the AGNN model and the meta-learning framework for indoor localization, incorporating data augmentation strategies and the distribution alignment method to address challenges posed by extremely sparse fingerprints in an increasingly connected world.

    \item We have collected several real datasets, utilizing a range of equipment from consumer-grade WiFi routers to specialized signal processing devices with trials conducted in both an open hall and a complex laboratory, yielding diverse data for fair performance evaluations. Extensive experiments on synthetic and real datasets demonstrate the AGML model’s superior performance in indoor scenarios with extremely sparse fingerprints, surpassing all baseline methods.
\end{itemize}

The remainder of this paper is organized as follows. 
Sec.~\ref {sec: bgd_proformu} presents the background of indoor localization and formulates a challenging localization problem with extremely sparse fingerprints. Sec.~\ref{sec: AGNN} introduces the AGNN model, and Sec.~\ref{sec: AGML} describes its integration into a meta-learning framework, followed by numerical results in Sec.~\ref{sec: experimental_results}. Finally, conclusions are drawn in Sec.~\ref{sec: conclusion}.

\textit{Notation:} Boldface lowercase letter $\bbx$ and boldface uppercase letter $\bbX$ are used to represent vectors and matrices, respectively.  $\bbx_i$ and $\bbx_{[i,:]}$ designate the $i$-th column and row of matrix $\bbX$, respectively. Meanwhile, the element in the $i$-th row and $j$-th column of matrix $\bbX$ is denoted by  $x_{ij}$. The calligraphic letter $\mathcal{S}$ is employed to denote a set, and $|\mathcal{S}|$ stands for the cardinality of this set. $[N]$ represents the set of natural numbers from $1$ to $N$. The operator $(\cdot)^\top$ denotes vector/matrix transpose. $\|\cdot\|$ represents the Euclidean norm of a vector. $\|\cdot\|_F$ denotes the Frobenius norm of a matrix.  $[\cdot\|\cdot]$ stands for the column/row concatenation operation for two column/row vectors.

\section{Background and Problem Formulation}
\label{sec: bgd_proformu}
This section first presents the localization scenario, highlighting the persistent challenge of labeled fingerprint collection. To address this challenge, we introduce two fingerprint augmentation strategies: unlabeled fingerprint augmentation and synthetic labeled fingerprint augmentation. Subsequently, we formally define the localization problem, along with real-world and synthetic datasets corresponding to our proposed augmentation strategies.

\subsection{Localization Scenario}
\label{sec: background}

We consider a challenging fingerprint-based indoor localization scenario characterized by various obstacles that impede signal propagation. Moreover, only a very limited number of strategically positioned labeled fingerprints are available, each accompanied by accurate spatial coordinates. 
We aim to determine the TP positions accurately in such a challenging scenario.

Several APs are strategically distributed throughout the area, playing a crucial role during the data collection phase. Both fingerprints and TPs are equipped with transmitters capable of emitting real-time signals, which are subsequently captured and recorded by the APs. The collected data encompasses a variety of metrics, including RSS, CSI, and CIR, among others, thereby providing a comprehensive characterization of the indoor environment.

However, the process of collecting a sufficiently diverse and labeled fingerprint dataset is inherently time-consuming and labor-intensive. To mitigate this challenge without compromising localization accuracy, innovative methodologies are imperative to augment existing indoor scene information. In this work, we propose two distinct strategies for augmenting fingerprint datasets: 1) unlabeled fingerprint augmentation and 2) synthetic labeled fingerprint augmentation, as illustrated in Fig.~\ref{fig: Indoor_Loc_Scenario}.

\vspace{4pt}
\textbf{\textit{Unlabeled Fingerprint Augmentation.}} 
This approach involves deploying mobile platforms equipped with transmitting antennas. These platforms traverse the indoor environment in a random manner, emitting signals from various locations. The resulting signals, captured and recorded by APs, provide a substantial repository of cost-effective, yet unlabeled, fingerprints, represented by the green dots in Fig.~\ref{fig: Indoor_Loc_Scenario} (a). 
These unlabeled fingerprints augment the sparse labeled fingerprints collected from predefined RPs, significantly enriching the real-world fingerprint database. By employing semi-supervised GNN architectures, this data augmentation strategy enables the model to incorporate information from unlabeled fingerprints during training, effectively mitigating the model underfitting issues that arise when training solely on the sparse labeled fingerprints.

\vspace{4pt}
\textbf{\textit{Synthetic Labeled Fingerprint Augmentation.}} 
This method utilizes advanced wireless signal simulators, such as Wireless Insite~\cite{6240749} or QuaDRiGa~\cite{6758357}, to simulate the target real indoor environment. This simulated indoor environment is meticulously modeled on the real-world scenario by considering various spatial attributes of obstacles, including their shapes, sizes, and distributions.
This process involves utilizing fingerprints and APs positioned at the same coordinates as in real-world settings. Additionally, a substantial number of fingerprints are deployed in the remaining areas to create a dense fingerprint field, as illustrated in Fig.~\ref{fig: Indoor_Loc_Scenario} (b). The simulator models the wireless signal transmission from fingerprints to APs, generating a synthetic labeled fingerprint dataset. 
To improve the model’s robustness and generalizability across similar scenarios, we introduce slight intentional perturbations in the simulation, such as small shifts in obstacle positions or variations in transmitter and receiver orientations. These perturbations produce multiple versions of the synthetic fingerprint datasets, thereby providing a diverse representation of possible environmental conditions.

\vspace{4pt}
Notably, both the unlabeled fingerprint augmentation and the synthetic labeled fingerprint augmentation require minimal human effort among others, thereby improving the overall effectiveness of fingerprint-based indoor localization systems.

\subsection{Problem Formulation}
\label{sec: problem_formulation}

We restrict our study to indoor localization in a two-dimensional (2-D) space, as the extension to the 3-D case is straightforward. According to the description in Sec.~\ref{sec: background}, the fingerprint dataset can be divided into two categories: the real-world fingerprint dataset and synthetic fingerprint datasets, denoted as $\ccalD_0$ and $\{\ccalD^1,\ccalD^2, \dots, \ccalD^m\}$, respectively.

\vspace{4pt}
\textbf{\textit{Real-world Dataset.}} For the dataset $\ccalD^0$, we define $\mathcal{S}_l = \{1,2,\ldots, N_l\}$ as the set of indices for labeled fingerprints, whose positions $\{\mathbf{p}_i\in\mathbb{R}^2, \forall i\in \mathcal{S}_l\}$ are known and fixed. Additionally, $\mathcal{S}_u^0 = \{N_{l+1}, N_{l+2},\ldots, N^0\}$ represents the set of indices for unlabeled fingerprints collected by mobile platforms. Each fingerprint is represented by a feature vector of dimension $F$.
Hence, we have $\ccalD^0:=\{\bbX^0\in\mbR^{N^0\times F};\bbY^0\in\mbR^{N_l\times 2}\}$. The unlabeled TPs are denoted by $\ccalD^t:=\{\bbX^t\in\mbR^{N^t\times F}\}$, where $N^t$ indicates the number of TPs.

\vspace{4pt}
\textbf{\textit{Synthetic Datasets.}} For $\{\ccalD^1,\ccalD^2, \dots, \ccalD^m\}$, there exists a set of indices for labeled fingerprints with identical positions, denoted as $\mathcal{S}_l$, while the set composed of additionally generated labeled fingerprints is represented as $\mathcal{S}_s^i = \{N_l+1, N_l+2,\ldots, N^i\}$. 
Thus, we define $\ccalD^i:=\{\mathcal{S}_l,\mathcal{S}_s^i\}=\{\bbX^i\in\mbR^{N^i\times F};\bbY^0\in\mbR^{N^i\times 2}\}$. Notably, in synthetic datasets, the number of labeled fingerprints is considerably greater than in real-world scenarios, specifically $N^i > N^0$. This increased density of labeled fingerprints is intended to ensure a more comprehensive representation of the environmental characteristics across the entire area.

\vspace{4pt}
Based on the aforementioned formulations, our primary objective is to train a model utilizing both real-world fingerprint datasets, $\ccalD^0$, and synthetic fingerprint datasets, $\{\ccalD^1,\ccalD^2, \dots, \ccalD^m\}$, to precisely locate all unlabeled TPs, $\ccalD^t$, while ensuring satisfactory computation time. 
To achieve this, we first introduce an Attentional GNN (AGNN) model, which acts as a foundation for our proposed method in this paper, to perform indoor localization using only the real dataset $\ccalD^0$ in the subsequent section.

\section{Attentional Graph Neural Network}
\label{sec: AGNN}

In this section, we adapt the AGNN~\cite{yan2024attentional}, originally designed for network localization based on inter-node distance measurements, to address fingerprint-based indoor localization using the real-world dataset $\ccalD^0$. The AGNN consists of three main components: the Distance Learning Module (DLM), the Adjacency Learning Module (ALM), and Multiple Graph Attention Layers (MGALs). Each of these components is discussed in detail below.

\subsection{Overview of AGNN Model}
\label{sec: overview_AGNN}

We formally define an undirected graph ${\mathcal{G}} = ({\mathcal{V}}, \mathbf{A})$, where $\mathcal{V}$ represents the set of nodes $\{v_1, v_2, \dots, v_{N^0}\}$, which correspond to both labeled and unlabeled fingerprints. The adjacency matrix $\mathbf{A} \in \mathbb{R}^{N^0 \times N^0}$ is a symmetric binary matrix.

In classical GNNs, such as the GCN \cite{kipf2016semi} and Graph Attention Network (GAT) \cite{velivckovic2017graph}, both the adjacency matrix and features of all nodes are provided. These models leverage local connectivity patterns encoded in the graph structure to capture spatial dependencies between nodes, enabling tasks such as node classification or regression. In fingerprint-based indoor localization, the dataset $\ccalD^0$ is typically available; however, the absence of an explicit adjacency matrix poses a significant challenge. This absence of structural information complicates the task of learning spatial correlations among fingerprints directly from $\ccalD^0$, which is critical for accurate localization. Therefore, the main challenge in such scenarios is how to effectively utilize the available data to infer spatial correlations between fingerprints and to predict their positions.

\begin{figure*}
    \centering
    \includegraphics[width=1\linewidth]{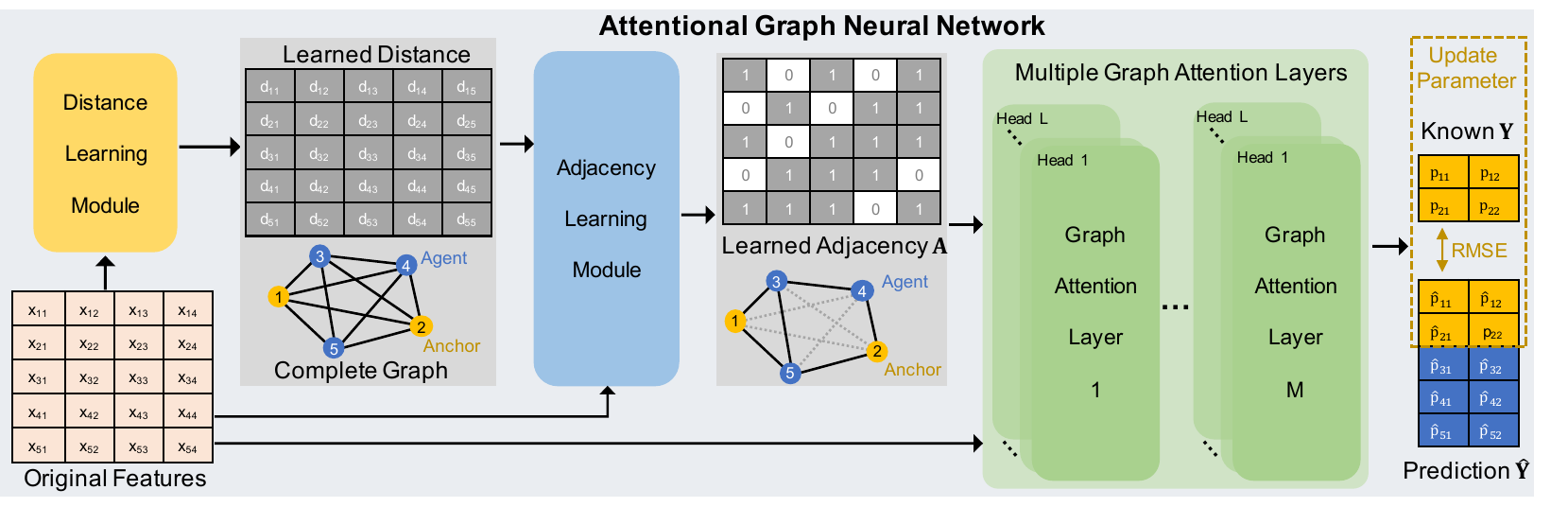}
    \caption{Model architecture of the tailored AGNN for indoor localization, which is composed of a DLM,  an ALM, and MGALs.}
    \label{fig: AGNN_Framework}
\end{figure*}

We previously proposed the AGNN model~\cite{yan2024attentional}, which consists of an ALM and MGALs to learn the adjacency matrix and achieve accurate network localization using inter-node distance measurements. However, in fingerprint-based indoor localization, such distance measurements between fingerprints are not available. Instead, the dataset $\ccalD^0$ contains only fingerprint information recorded by a limited set of APs, represented by CSI or CIR data corresponding to each RP. 

To overcome this limitation, we tailored the original AGNN to address the specific requirements of fingerprint-based indoor localization by incorporating an additional DLM.
The architecture of the tailored AGNN for indoor localization is depicted in Fig.~\ref{fig: AGNN_Framework}. 
The DLM employs a Multi-Layer Perceptron (MLP) to transform the high-dimensional CIR or CSI features into a compact, lower-dimensional representation. This transformation is followed by the computation of the Euclidean norm between node embeddings to establish a metric for node proximity.
The ALM leverages an attention mechanism to learn a proximity-aware threshold for adjacency matrix construction. In essence, this enables nodes to adaptively adjust their threshold based on the proximity to neighboring nodes, facilitating a flexible selection of neighbors. 
Following this, the MGALs utilize another attention mechanism to perform node localization, utilizing the adjacency matrix inferred by the ALM.

\subsection{Distance Learning Module (DLM)}
\label{sec: distance_learning_module}

To construct adjacency relationships, we first compute the proximity between each pair of nodes. Specifically, the proximity metric between nodes $v_i$ and $v_j$ is defined as:
\begin{equation}
	d_{ij} = \|\bbl_i-\bbl_j\|_2,
\end{equation}
where $\mathbf{l}_i$ and $\mathbf{l}_j$ are the latent embeddings of nodes $v_i$ and $v_j$, respectively. These embeddings are obtained by transforming the original feature space using the following equation:
\begin{equation}
	\label{eq: feature_transformation}
	\bbL=\sigma(\bbX^0\bbM^1)\bbM^2.
\end{equation}
In this equation, $\bbM^1\in\mbR^{F\times D_{M_1}}$ and $\bbM^2\in\mbR^{D_{M_1}\times D_{M_2}}$ are trainable weight parameter matrices for the proximity computation, typically with $D_{M_1}\approx D_{M_2}<F$,  
and $\sigma(\cdot)$ is an activation function for introducing nonlinearity into the feature mapping. The resulting matrix $\bbL$  contains the fingerprint embeddings of nodes $v_i$ and $v_j$ in the latent space. For the localization task, the Euclidean distance $d_{ij}$ between these embeddings quantifies the closeness of the fingerprints within the latent space. This proximity metric is crucial for determining node adjacency in the graph, a key aspect for precise indoor localization. The utilization of this metric will be detailed in the following section.

\subsection{Adjacency Learning Module (ALM)}
\label{sec: attentional_neighbor_selection}

Once the proximity metric is computed, we determine the adjacency of each pair of nodes by comparing their proximity to a specific threshold.
To adaptively learn this proximity-aware threshold during training, we incorporate an attention mechanism to determine a specific threshold, $T^{A}_{ij}$, for each node pair. Nevertheless, learning $T^{A}_{ij}$ directly for each pair without prior graph structure knowledge can lead to high computational complexity and over-smoothing issues \cite{yan2024attentional}. To mitigate these drawbacks, the ALM employs a two-stage process:  a coarse-grained neighbor selection followed by fine-grained neighbor refinement using attention scores.

\vspace{4pt}
\textbf{\textit{Stage-I: Coarse-grained Neighbor Selection.}}
To address the challenge of computing attention scores without prior knowledge of the graph structure, we perform coarse-grained neighbor selection using a manually set threshold $T_h^0$. The procedure for selecting a coarse neighborhood is as follows:
\begin{equation}
    \mathcal{N}_i^C = \left\{j\;|\;d_{ij} <T_h^0\right\}.
\end{equation}
This initial threshold $T_h^0$ can be set manually or optimized, as elaborated in detail in our previous AGNN model for network localization \cite{yan2024attentional}. Notably, this initial threshold can be chosen from a broad range, as our objective here is to establish a coarse-grained neighbor set, $\mathcal{N}_i^C$, for each node $i$.

\vspace{4pt}
\textbf{\textit{Stage-II: Fine-grained Neighbor Refinement.}}
We refine the coarse-grained neighbor into a fine-grained one by applying the learned proximity-aware threshold matrix via attention scores. 
Specifically, we calculate the attention scores $e_{ij}^A$ for nodes $j \in \mathcal{N}_i^C$ based on the masked attention mechanism \cite{vaswani2017attention}:
\begin{equation}
	\label{eq: neighbor_selection_att}
	e_{ij}^A =\left| \phi\left( \bbx^0_{[i,:]}\bbW_A \right) -   \phi\left(\bbx^0_{[j,:]}\bbW_A \right) \right| \bbv_A,\;  j\in \ccalN_i^C.
\end{equation}
Here, $\lvert\bbx\rvert$ takes the element-wise absolute value of a vector $\mathbf{x}$. The attention mechanism is parametrized by an attention weight vector $\bbv_A \in \mathbb{R}^{F_A}$ and an attention weight matrix $\bbW_{A}\in \mathbb{R}^{F \times F_A}$, and applies a nonlinear mapping $\phi(\cdot)$, such as LeakyReLU.
The rationale behind Eq.~\eqref{eq: neighbor_selection_att} can be explained as follows: 
\begin{itemize}
    \item Applying a linear transformation with $\bbW_A$ followed by a nonlinear activation, $\phi(\cdot)$, enhances the expressiveness of feature transformations; 
    \item Taking the absolute difference between the transformed representations captures a learnable relevance metric that inherently maintains symmetry, a factor not considered in previous works but essential for indoor localization.
    \item The attention vector $\bbv_A$ projects this relevance metric into a scalar, producing the final attention score.
\end{itemize}

To convert the attention scores into thresholds, we apply a nonlinear scaling function:
\begin{equation}
	\label{eq: scale_att}
	T_{ij}^A  = \mathrm{max}(\bbd_{[i,:]})\cdot\mathrm{Sigmoid}({e}_{ij}^A),
\end{equation}
where $\max(\mathbf{d}_{[i,:]})$ is the maximum value in the row vector $\bbd_{[i,:]}$. The resulting proximity-aware threshold matrix $\bbT^A\in\mathbb{R}^{N\times N}$ is used to adjust the adjacency relationships between nodes.

Finally, the adjacency matrix is constructed based on a comparison between the proximity metric and the learned threshold: 
\begin{equation}\label{eq:thre_adj_att}
	a_{ij}=\begin{cases}
		1 ,\quad \mathrm{if} \quad  d_{ij} <T_{ij}^A, \\
		0,\quad \mathrm{otherwise}.
	\end{cases}
\end{equation}

To ensure differentiability with respect to the learnable parameters, we approximate the step function using a more suitable activation function: 
\begin{equation}
\hat{\delta}(x) = \mathrm{ReLU}(\mathrm{tanh}(\gamma x)), 
\end{equation} 
where $\gamma$ is a hyperparameter that adjusts the steepness of the transition\footnote{The value of $\gamma$ is typically determined based on the magnitude of the input $x$, such that the product $\gamma x$ remains within the interval $[-10,10]$. This ensures that the transition of the approximated step function is both smooth and sufficiently sharp for learning.}. The devised approximated step function, employing ReLU and hyperbolic tangent (tanh), yields a notably sharper transition on the positive side and effectively truncates negative values to zero, demonstrating an accurate approximation of the step function. More importantly, its derivative can be obtained everywhere with respect to the trainable threshold. For more details, see 
\ifappendixincluded
App.~\ref{app: appro_step_func}.
\else
Appendix F in the supplementary material.
\fi

The final adjacency matrix $\mathbf{A}$ is then derived by applying this approximation:
\begin{align}
	a_{ij} = \mathrm{ReLU}(-\mathrm{tanh}(\gamma(d_{ij}-T_{ij}^A) )).
\end{align}
This formulation yields the adjacency matrix $\bbA$, which is subsequently utilized as the input of the MGALs introduced below.

\vspace{4pt}
\subsection{Multiple Graph Attention Layers (MGALs)}
\label{sec: graph_attention_layer}

Once the fine-grained neighbor set $\mathcal{N}_i^F$ for each node $i$ is obtained, we apply MGALs, a more generalized version of the graph attention mechanism inspired by GATv2 \cite{brody2022how}, to learn the aggregation weight and predict the positions of all nodes. We begin by elaborating the structure and operations of the $k$-th individual graph attention layer, which serve as the foundation of MGALs.

The node representation at the $k$-th layer is denoted as $\mathbf{H}^{(k)} \in \mathbb{R}^{N \times D_k}$, where $D_k$ represents the hidden dimension of the $k$-th layer. The initial node representation is set to be the feature matrix, i.e., $\mathbf{H}^{(1)} = \mathbf{X}^0$. In each graph attention layer, we employ a single-layer feed-forward neural network parameterized by an attention weight vector $\bbv_{att}^{(k)} \in \mathbb{R}^{F_{att}}$ and an attention weight matrix $\bbW_{att}^{(k)} \in \mathbb{R}^{2D_{k+1} \times F_{att}}$ to implement the attention mechanism. The attention scores for node $i$ and its neighboring nodes $j \in \mathcal{N}_i^F$ are computed as follows:
\begin{equation}
    \label{eq: dynamic_GAT_att}
e_{ij}^{(k)} =\phi\left( \left[ {\hbh}_{[i,:]}^{(k)} \left\| {\hbh}_{[j,:]}^{(k)}\right.  \right]  \bbW_{att}^{(k)}  \right)  \bbv_{att}^{(k)},
\end{equation}
where ${\hbh}_{[i,:]}^{(k)}$ and ${\hbh}_{[j,:]}^{(k)}$ are transformed node feature vectors for nodes $i$ and $j$, respectively, defined as follows:
\begin{align}
\label{eq: feature_trans}
\hbh_{[i,:]}^{(k)} =\bbh_{[i,:]}^{(k)} \bbW^{(k)}, \quad
\hbh_{[j,:]}^{(k)} =\bbh_{[j,:]}^{(k)} \bbW^{(k)}.
\end{align}  
Here, ${\bbW^{(k)} }\in \mbR^{D_{k}\times D_{k+1}}$ represents the weight matrix in the ${k}$-th graph attention layer. 

Notably, two distinct learnable weight matrices are used in our approach: $\bbW^{(k)}$ and $\bbW_{att}^{(k)}$. The matrix $\bbW^{(k)}$ is responsible for the linear transformation of node features, while $\bbW_{att}^{(k)}$ and the attention weight vector $\bbv_{att}^{(k)}$ are used to compute the attention scores between node pairs, reflecting the relevance of each neighbor.
It is noteworthy that this attention mechanism presents a more generalized form than GATv2 and can be reduced to GATv2 when $F_{att}^{k}=2D_{{k+1}}$ and $\bbW_{att}^{(k)}$ assumes an identity matrix.

To make the attention scores comparable across different nodes, we apply the $\mathrm{softmax}$ function to normalize the scores over the neighbors of each node $i$:
\begin{equation}
	\label{eq: softmax_GAT_att}	
	\alpha_{ij}^{(k)} = \frac{\exp({e}_{ij}^{(k)} )}{\sum_{j'\in\ccalN_i^F} \exp({e}_{ij'}^{(k)} )}.
\end{equation}
The normalized attention scores $\alpha_{ij}^{(k)}$ are then used to update the representation of node $i$ in the $(k+1)$-th layer, with the aid of a non-linear activation function $\varphi(\cdot)$:
\begin{equation}
\label{eq: dynamic_GAT_eqnatt}
	\bbh^{(k+1)}_{[i,:]} = \varphi\left(\sum_{j\in\ccalN_i^F} \alpha_{ij}^{(k)} \hbh_{[j,:]}^{(k)} \right).
\end{equation}
After passing through the MGALs, the output of the final layer represents the predicted position matrix, denoted as $\bbH^{(K)} = \hbY$.

In summary, we propose a DLM that learns a proximity matrix and introduces an ALM capable of determining optimal thresholds for individual edges through a novel attention mechanism. This ALM further facilitates the learning of an adaptive adjacency matrix. MGALs then leverage these learned matrices and apply another attention mechanism to acquire adaptive aggregation weights, significantly enhancing the expressiveness of the model.

Finally, we formulate the optimization problem for the proposed AGNN model as follows:
\setlength{\jot}{1pt}
\begin{equation}
		\begin{aligned}
		\mathop{\arg\min}\limits_{\boldsymbol{\theta}}~ \mathcal{L}\left( f_{\boldsymbol{\theta}};\ccalD^0\right)  :&= \|\mathbf{Y}_l-\hat{\mathbf{Y}}_l \|_F\\[2\jot]
		\mathrm{s.t.} ~~~~~~~~~~~~~~ \hat{\bbY} &= \text{MGALs}(\bbA,{\bbX^0})\\[\jot]
		\mathbf{A} &= \text{ALM}(\bbX^0, \bbD, T_h^{0})\\[\jot]
		\bbD &=\text{DLM}(\bbX^0),\\[\jot]
	\end{aligned}
\end{equation}
where the objective function represents the Frobenius norm of the difference between the true positions of labeled fingerprints, $\mathbf{Y}_l = [\mathbf{y}_1, \mathbf{y}_2, \dots, \mathbf{y}_{N_l}]^{\top}$, and their corresponding estimates, $\hat{\mathbf{Y}}_l = [\hat{\mathbf{y}}_1, \hat{\mathbf{y}}_2, \dots, \hat{\mathbf{y}}_{N_l}]^{\top}$. Here, $f_{\boldsymbol{\theta}}$ denotes the AGNN model, and $\boldsymbol{\theta}$ represents all trainable parameters within the AGNN framework, including those in the DLM, ALM, and MGALs components. The optimization problem can be solved using gradient-based optimization methods, such as those described in \cite{bottou2010large} and \cite{kingma2014adam}.

\section{Attentional Graph Meta-learning}
\label{sec: AGML}

In this section, we first introduce the meta-learning framework, specifically Model-Agnostic Meta-Learning (MAML), that enables a model to quickly adapt to new tasks with minimal additional training. Subsequently, we introduce a novel model termed Attentional Graph Meta-Learning (AGML), which is capable of leveraging data from both the real-world fingerprint dataset $\ccalD^0$ and some synthetic fingerprint datasets $\{\ccalD^1,\ccalD^2, \dots, \ccalD^m\}$, to achieve accurate localization of users for harsh indoor environments.

\subsection{Meta-Learning}

Meta-learning is a learning-to-learn approach that enables the learning model to adapt to new tasks by leveraging previous experience from related tasks. In this framework, tasks are sampled from a specific distribution, denoted as $\ccalT\sim\mathcal{P}(\ccalT)$, and each task includes a support set for training and a query set for testing. In an $N$-way $k$-shot classification problem, a task consists of $N$ classes, each with $k$ samples. In the meta-training stage, $M$ training tasks, $\{\ccalT_{i}\}_{i=1}^{M}\sim\mathcal{P}(\ccalT)$, are sampled from the distribution and the corresponding datasets are made available to the model. In the meta-test stage, a new test task $\ccalT_t\sim\mathcal{P}(\ccalT)$ is presented, consisting of a small support set and a query set. The objective of meta-learning is to train a model on the $M$ training tasks, such that it can quickly adapt to the new test task using a support set of small size and perform well on the query set. 

MAML is able to achieve this by learning a set of initial parameters $\boldsymbol{\theta}_\text{MAML}$ for a neural network type of architecture that enables good performance on a new task with only a few gradient descent steps. In the meta-training stage, MAML formulates a meta-optimization problem to find $\boldsymbol{\theta}_\text{MAML}$ as:
\begin{equation}    \boldsymbol{\theta}_\text{MAML}=\mathop{\arg\min}\limits_{\boldsymbol{\theta}}\sum_{i=1}^{M}\hat{\mathcal{L}}_{\ccalT_{i}}\left(\boldsymbol{\theta}-\alpha\nabla_{\boldsymbol{\theta}}{\mathcal{L}}_{\ccalT_{i}}(\boldsymbol{\theta})\right),
\end{equation}
where it contains two task-specific loss functions ${\mathcal{L}}_{\ccalT_{i}}$ and
$\hat{\mathcal{L}}_{\ccalT_{i}}$ computed based on the support set and query set of the training task $\ccalT_{i}$, respectively. Then the meta-parameters are updated via Stochastic Gradient Descent (SGD):
\begin{equation}
\boldsymbol{\theta}_\text{MAML}\gets\boldsymbol{\theta}_\text{MAML}-\beta\nabla_{\boldsymbol{\theta}}\sum_{i=1}^{M}\hat{\mathcal{L}}_{\ccalT_{i}}\left(\boldsymbol{\theta}-\alpha\nabla_{\boldsymbol{\theta}}{\mathcal{L}}_{\ccalT_{i}}(\boldsymbol{\theta})\right),
	\label{eq:outer_loop}
\end{equation}
where $\alpha$ and $\beta$ denote the step size of the inner loop and outer loop, respectively. During the meta-test stage, the meta-parameters $\boldsymbol{\theta}_\text{MAML}$ are fine-tuned to obtain the parameters $\boldsymbol{\theta}_{\ccalT_t}$ for the neural network to be used in the test task $\ccalT_t$. This is achieved by updating the meta-parameters using the gradient of the loss function ${\mathcal{L}}_{\ccalT_t}\left(\boldsymbol{\theta}_\text{MAML}\right)$ computed based on the support set of the test task, namely:
\begin{equation}
	\boldsymbol{\theta}_{\ccalT_t}\gets\boldsymbol{\theta}_\text{MAML}-\alpha\nabla_{\boldsymbol{\theta}}{\mathcal{L}}_{\ccalT_t}\left(\boldsymbol{\theta}_\text{MAML}\right).
\end{equation}

\subsection{AGML Model for Fingerprint-based Indoor localization}
\label{sec: AGML_steps}

\begin{figure*}[t] 
\centering
\includegraphics[width=0.9\linewidth]{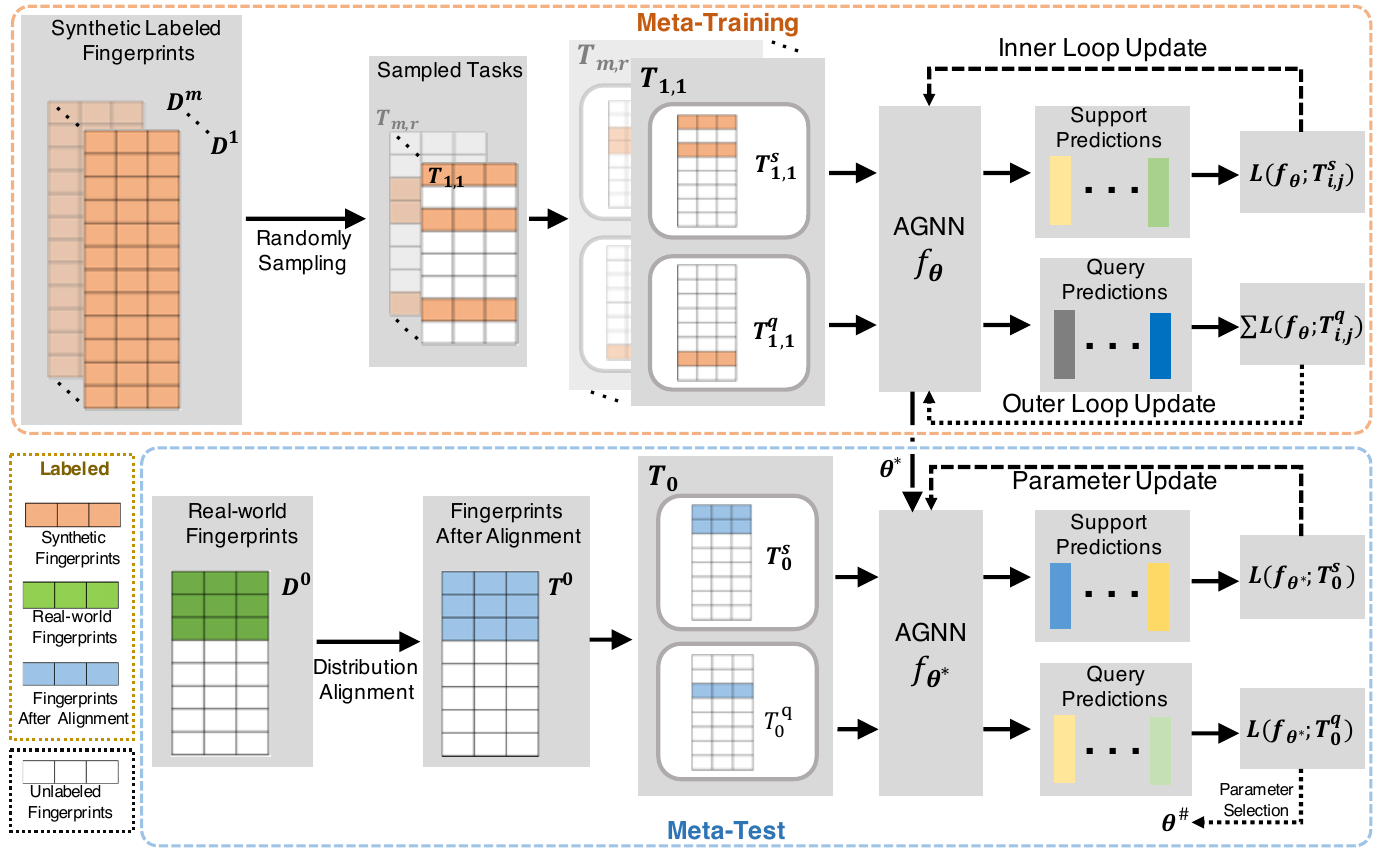}
\caption{The framework of the AGML model, which consists of a meta-training phase on synthetic datasets and a meta-test phase on the real-world dataset. In meta-training, three steps are performed: task definition ($m$ synthetic datasets are randomly sampled into $m\times r$ tasks), inner loop update (the model performs one-step gradient descent on each task's support set), and outer loop update (meta-parameters are optimized across multiple tasks' query sets). In meta-testing, the model undergoes: distribution alignment (aligning the feature distributions of the real-world dataset with synthetic datasets), model adaptation (updating the meta-parameters using the transformed dataset), and localization of TPs (locating all unlabeled TPs).}
\label{fig: AttGraph_MetaLearning}
\end{figure*}

Motivated by the promising results achieved through meta-learning methodologies, we propose the AGML model, which combines the strengths of the AGNN model and the meta-learning framework. AGML leverages both the real-world fingerprint dataset $\ccalD^0$ and the synthetic dataset $\{\ccalD^1,\ccalD^2, \dots, \ccalD^m\}$ to achieve accurate user localization. By leveraging the complementary information provided by both datasets, AGML aims to enhance localization performance by learning robust representations of indoor environments and effectively capturing the spatial dependencies between fingerprints. The framework of the AGML model is depicted in Fig.~\ref{fig: AttGraph_MetaLearning}.

Specifically, this process comprises two distinct stages: the meta-training phase and the meta-test phase, both of which share the same set of parameters in the AGNN model, maintaining identical architectural configurations throughout. The meta-training phase takes place with the simulated environments, where a substantial number of localization tasks are extracted from densely sampled fingerprint datasets using the AGNN model for offline training. This endeavor aims to acquire meta-parameters for the trained model. These meta-parameters are subsequently utilized as the initial parameters for the AGNN in the meta-test phase, which is conducted in real-world scenarios. Their details are as follows.

\subsubsection{Meta-training Phase}
The primary purpose of the meta-training stage is to derive a well-trained meta-parameter set $\boldsymbol{\theta}^*$. The process is carried out in the following steps:

\vspace{4pt}
\textbf{\textit{Step-I: Task Definition.}} For each synthetic dataset $\ccalD^i$, which contains $N^i$ labeled fingerprints, we sample $N^0$ fingerprint points based on certain rules (e.g., mimicking fingerprints' distribution density in $\ccalD^0$), forming a task. Each task consists of $N_l$ labeled and $(N^0-N_l)$ unlabeled fingerprints. By sampling with replacement $r$ times from $\ccalD^i$, we generate a set of tasks $\{\ccalT_{i,1}, \ccalT_{i,2}, \dots, \ccalT_{i,r}\}$. Iterating over all $m$ datasets $\{\ccalD^1, \ccalD^2, \dots, \ccalD^m\}$ yields a comprehensive collection of tasks $\{\ccalT_{1,1}, \dots, \ccalT_{m,r}\}$. This strategy promotes the adaptability of our localization system in diverse environments through extensive data sampling.

Each task $\ccalT_{i,j}$ consists of a support set $\ccalT_{i,j}^{s}$ and a query set $\ccalT_{i,j}^{q}$. Both sets contain feature representations for all $N^0$ fingerprints, with the distinction being the proportion of labeled fingerprints. Formally, we define the support set as $\ccalT_{i,j}^{s}:= \{\bbX^{i,j} \in \mathbb{R}^{N^0 \times F}, \bbY^{i,j}_s \in \mathbb{R}^{(p \cdot N_l) \times 2} \}$, and the query set as $\ccalT_{i,j}^{q}:= \{\bbX^{i,j} \in \mathbb{R}^{N^0 \times F}, \bbY^{i,j}_q \in \mathbb{R}^{((1-p) \cdot N_l) \times 2} \}$, where $p$ denotes the proportion of labeled data in the support set.

During meta-training, a dual-layer mechanism updates the parameter set $\boldsymbol{\theta}$ of AGNN. Differing from traditional single-layer gradient updates, meta-learning seeks an optimal $\boldsymbol{\theta}^*$ by learning from multiple tasks, enabling AGNN to quickly adapt to new tasks. This mechanism involves two loops: the inner loop and the outer loop.

\vspace{4pt}
\textbf{\textit{Step-II: Inner Loop Update.}} The inner loop operates within each task to facilitate rapid adaptation. For each task $\ccalT_{i,j}$, the loss function ${\mathcal{L}}(f_{\boldsymbol{\theta}})$ is computed on the support set $\ccalT_{i,j}^{s}$.
Model parameters are then updated using a single gradient descent step to minimize the loss:
\begin{equation}
    {\boldsymbol{\theta}}^\prime_{i,j} = {\boldsymbol{\theta}}-\alpha\nabla_{\boldsymbol{\theta}} {\mathcal{L}}(f_{{\boldsymbol{\theta}}};\ccalT_{i,j}^{s}),
\end{equation}
where $\alpha$ denotes the learning rate of the inner loop, and $\nabla_{\boldsymbol{\theta}}\ccalL$ represents the gradient of the loss function with respect to the model parameters. 
The task-specific parameters ${\boldsymbol{\theta}}^\prime_{i,j}$ provide only limited information about each localization task, as they are derived from a single gradient update on the support set. Further evaluation on the query set $\ccalT_{i,j}^{q}$ is needed for a more comprehensive understanding of each task's performance.

\vspace{4pt}
\textbf{\textit{Step-III: Outer Loop Update.}} The outer loop aims to update the meta-parameters for generalization across multiple tasks. 
Unlike the inner loop, which updates task-specific parameters, the outer loop optimizes for broader learning across tasks. For each query set $\ccalT_{i,j}^{q}$, the loss $\ccalL(f_{\boldsymbol{\theta}^\prime_{i,j}}; \ccalT_{i,j}^{q})$ is computed using the task-specific parameters $\boldsymbol{\theta}^\prime_{i,j}$. The meta-loss is the sum of these losses across tasks, and the optimal meta-parameters are determined by:
\begin{equation}
    \boldsymbol{\theta}^* = \mathop{\arg\min}\limits_{\boldsymbol{\theta}}
    \sum_{i,j=1}^{m,r}\ccalL(f_{\boldsymbol{\theta}^\prime_{i,j}};\ccalT_{i,j}^q).
\end{equation}
The meta-parameters are updated via gradient descent:
\begin{equation}
    \boldsymbol{\theta}^* \leftarrow \boldsymbol{\theta}-\beta\nabla_{\boldsymbol{\theta}} \sum_{i,j=1}^{m,r}\ccalL(f_{\boldsymbol{\theta}^\prime_{i,j}};\ccalT_{i,j}^q).
\end{equation}
Here, $\beta$ denotes the learning rate of the outer loop. 
The outer loop ensures that the model not only performs well on individual tasks but also effectively learns and generalizes across a series of tasks.

The outer loop plays a critical role in refining meta-parameters based on the aggregate performance of the model across multiple tasks. Through iterative optimization, the model learns to generalize effectively, ensuring adaptability and strong performance across diverse tasks encountered during meta-testing.

\subsubsection{Meta-test Phase}
The primary purpose of the meta-test stage is to update the meta-parameters $\boldsymbol{\theta}^*$ acquired during the meta-training phase for the real-world fingerprint dataset $\ccalD^0$, ultimately facilitating the accurate localization of all unlabeled TPs $\ccalD^t$. This phase unfolds in a series of systematic steps:

\vspace{4pt}
\textbf{\textit{Step-IV: Distribution Alignment.}}
The primary objective of the distribution alignment phase is to align the feature distributions of the real-world fingerprint dataset $\ccalD^0$ with those of the synthetic datasets $\{\ccalD^1, \ccalD^2, \dots, \ccalD^m\}$.

This approach is based on the premise that synthetic data is generated using models that replicate real-world environments, thereby adhering to similar physical principles. For example, both real and synthetic data exhibit an increase in signal strength and a decrease in time of arrival when the transmitter and receiver stay close to each other. Although synthetic data may show slight variations in signal amplitude due to the assumptions and simplifications made during the simulation process, the fundamental relationships between features remain consistent with those observed in real-world data.

To facilitate alignment between the distributions, we fit basic statistical models to each feature dimension $j, \forall j\in\{1,2,\dots, F\}$, for both the real-world dataset $\ccalD^0$ and the synthetic datasets $\{\ccalD^1, \ccalD^2, \dots, \ccalD^m\}$. This modeling enables us to derive key statistics for each feature, such as the mean and standard deviation, denoted by $\{\mu_j^0, \sigma_j^0\}$ for the real-world dataset and $\{\mu_j^s, \sigma_j^s\}$ for the synthetic datasets.

Subsequently, a transformation is applied to the real-world dataset $\ccalD^0$ to align it with the statistical properties of the synthetic datasets, resulting in the transformed fingerprints $\ccalT^0$. This transformation standardizes the real-world features and adjusts them to match the distributions of the synthetic features. The transformation for each feature $j$ is given by: 
\begin{equation}
\hbX^0_j = \sigma^s_j \frac{\bbX^0_j - \mu_j^0}{\sigma^0_j} + \mu_j^s.
\end{equation} 
This process allows the real-world data to be adapted to the statistical distribution of the synthetic data while preserving the inherent relationships between features. After applying this transformation to each feature dimension, we obtain the transformed dataset $\ccalT^0:= \{\hbX^0 \in \mbR^{N^0 \times F}; \bbY^0 \in \mbR^{N_l \times 2}\}$. The support set of the transformed dataset is defined as $\ccalT_0^s := \{\hbX^0 \in \mbR^{N^0 \times F}; \bbY^0_s \in \mbR^{(p \cdot N_l) \times 2}\}$, and the query set as $\ccalT_0^q := \{\hbX^0 \in \mbR^{N^0 \times F}; \bbY^0_q \in \mbR^{((1-p) \cdot N_l) \times 2}\}$.

\vspace{4pt}
\textbf{\textit{Step-V: Model Adaptation.}}
The meta-parameters $\boldsymbol{\theta}^*$ derived from the meta-training phase serve as the initial parameters for the AGNN model. These parameters are subsequently fine-tuned on the support set $\ccalT^s_0$, yielding optimal parameters for the new environment:
\begin{equation}
    \boldsymbol{\theta}^\# = \mathop{\arg\min}\limits_{\boldsymbol{\theta}} \ccalL(f_{\boldsymbol{\theta}^*};\ccalT^s_0).
\end{equation}
Gradient descent is employed to update $\boldsymbol{\theta}^*$ and minimize the loss function, resulting in:
\begin{equation}
    \boldsymbol{\theta}^\# \leftarrow \boldsymbol{\theta}^*-\zeta\nabla_{\boldsymbol{\theta}} \ccalL(f_{\boldsymbol{\theta}^*};\ccalT^s_0),
\end{equation}
where $\zeta$ denotes the learning rate for the target environment. 
The query set of the transformed dataset $\ccalT_0^q$ functions similarly to a validation set commonly employed in neural network training, facilitating the selection of parameters $\boldsymbol{\theta}^\#$, that yield optimal performance on $\ccalT_0^q$ by minimizing $\ccalL(f_{\boldsymbol{\theta}^\#};\ccalT^q_0)$. This approach also helps mitigate the risk of overfitting the model to the transformed dataset $\ccalT^0$.

\vspace{4pt}
\textbf{\textit{Step-VI: Localization of TPs.}}
Following the completion of the model adaptation process, we proceed to predict the coordinates of all unlabeled TPs $\ccalD^t$. Initially, we apply the same distribution alignment methodology outlined in Step-IV to transform $\ccalD^t$ into the format $\ccalT^t:=\{\hbX^t\in\mbR^{N^t\times F}\}$. 

Next, leveraging the semi-supervised learning capabilities of the AGNN model, we concatenate the transformed feature matrices $\hbX^0$ and $\hbX^t$ along the column dimension to create a unified input for the AGNN. This concatenation allows the model to utilize the complementary information from both datasets effectively. The combined input is expressed as:
\begin{equation}
    [\hbY^0;\hbY^t]=f_{\boldsymbol{\theta}^\#}([\hbX^0;\hbX^t]),
\end{equation}
where $\hbY^t\in\mbR^{N^t\times 2}$ denotes the predicted coordinates for all unlabeled TPs. This approach facilitates accurate localization by harnessing the learned representations from the semi-supervised framework, ultimately enhancing the model's performance in predicting the locations of the TPs.

\section{Experimental Results}
\label{sec: experimental_results}
\subsection{Data Generation and Experimental Setup}
\label{sec: data_generation}

To evaluate the proposed method, we establish two primary types of datasets: real-world datasets obtained through site surveys, and synthetic datasets generated using a ray-tracing platform. This section provides a comprehensive overview of the real-world data collection process, including the details of various platforms used and the generation of synthetic data. These datasets serve as the foundation for our model validation and experimentation.

\subsubsection{Real‑world Data from Site Surveys}
The real-world dataset is collected through two distinct setups: one using commercial WiFi devices and another using high-precision professional equipment. Both setups allow for comprehensive channel measurements, capturing small-scale multipath effects and signal propagation characteristics.

\vspace{4pt}
\textbf{\textit{Commercial WiFi Platform.}}
For the WiFi setup, we leverage the Nexmon CSI platform on commercial WiFi devices, which provides an accessible and cost-effective means of capturing the frequency-domain CSI from 802.11ac WiFi frames. The detailed data collection setup for the WiFi platform is provided in 
\ifappendixincluded
App.~\ref{app: uplink-communication-framework}.
\else
Appendix B.
\fi
The commercial WiFi system, while convenient, introduces some challenges due to the limitations of the hardware.
Specifically, CSI obtained from commercial WiFi may include signal distortions caused by signal superposition during propagation and hardware signal processing issues, such as inaccurate sampling frequencies and carrier frequency offsets~\cite{CRISLoc,8423070}. These CSI errors will significantly impact the derived CIR and completely alter the multipath characteristics~\cite{8423070}. To mitigate these issues and ensure accurate fingerprinting, we employ several preprocessing steps, including abnormal packet removal, CSI amplitude calibration, CSI phase correction, and transformation from CSI to CIR, which are detailed in
\ifappendixincluded
App.~\ref{app: data_preprocess}. 
\else
Appendix A.
\fi

\vspace{4pt}
\textbf{\textit{Professional Communication Platform.}}
While the commercial WiFi platform is convenient, it suffers from hardware limitations that can introduce significant noise and measurement instability. To further validate our method, we utilize high-precision professional wireless channel measurement equipment. The detailed data collection setup for the professional communication platform is provided in 
\ifappendixincluded
App.~\ref{app: setups-keysight}.
\else
Appendix C.
\fi

\subsubsection{Synthetic Data from Computer Simulations}
In addition to real-world data, we generate synthetic datasets using the Wireless Insite (WI) ray-tracing platform, a tool developed by REMCOM that simulates and predicts complex scenarios using advanced electromagnetic methods. Specifically, WI launches a large number of rays from each transmitter at a certain density. Based on the defined environmental parameters, the platform computes the propagation loss of each ray. If the received power of a path exceeds a predefined threshold, the ray's power and time of arrival are recorded as a valid path; otherwise, the ray is discarded. An example of CIR generated from the WI platform can be found in 
\ifappendixincluded
App.~\ref{app: wi-cir}.
\else
Appendix D.
\fi

In the WI platform, we set the maximum number of reflections, transmissions, and diffractions to 4, 1, and 1, respectively~\cite{remcom_wifi_simulator}. Ray density is adjusted through ray spacing; a smaller spacing results in more densely emitted rays. In our experiment, we set the ray spacing to 0.2 degrees and limit the number of paths to 500. The received power threshold is set to -250 dBm to filter out low-power noise. We use the default isotropic antenna model provided by the simulation platform. 

\begin{figure}[t]
    \centering
    \includegraphics[scale=0.43]{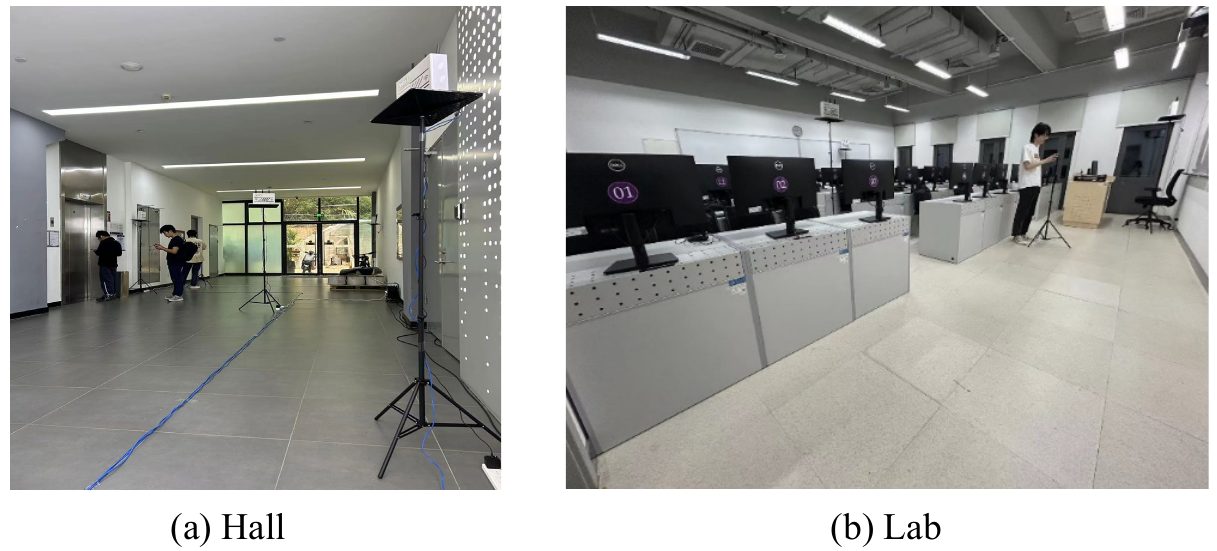}
    \caption{Photographs of the Hall-scenario and Lab-scenario on the CUHKSZ campus.}
    \label{fig:real}
\end{figure}

\begin{figure}[t]
    \centering
    \includegraphics[scale=0.4]{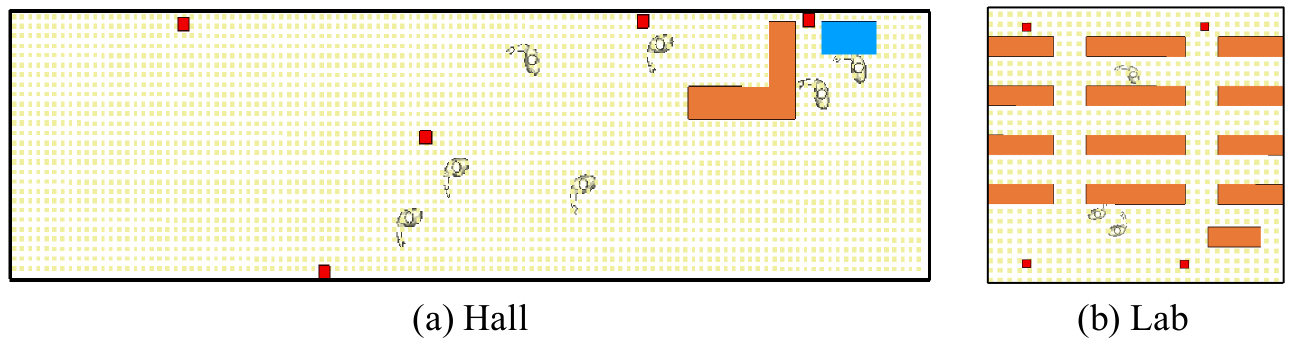}
    \caption{Simulations of the Hall-scenario and Lab-scenario on the CUHKSZ campus, which include the same uplink communication system and AP positions as the real world. The blue area represents clutter, the orange areas represent desks, the red dots indicate receiver nodes, and the yellow dots indicate transmitter nodes. Additionally, some people are included in the simulation to better mimic real-world conditions.}
    \label{fig:simulation}
\end{figure}

\subsubsection{Data Collection Setup} 
The data collection process is carried out in two distinct environments: the hall and the laboratory, as depicted in Fig.~\ref{fig:real} for the real-world environment and Fig.~\ref{fig:simulation} for the simulated environment, respectively. The hall represents a Line-of-Sight (LOS) dominated scenario, characterized by an open area with minimal obstructions. In contrast, the laboratory environment is configured as an NLOS scenario, where the presence of obstacles such as desks, computers, and other equipment altogether creates a more complex propagation environment with significant signal reflections and diffractions.
In each of these environments, several APs are strategically deployed to ensure comprehensive coverage and capture a diverse range of channel conditions. CSI or CIR data is systematically collected from a series of measurement points that are evenly distributed throughout the area. This approach ensures that the collected data  is representative of the entire environment, capturing variations in signal propagation due to different spatial configurations and obstacles. A summary of the data collection setup is presented in Tab.~\ref{tab: Data Collection Setup}.

In particular, the commercial WiFi platform initially captures data in the form of frequency-domain CSI. 
This raw CSI data is preprocessed (abnormal packet removal, CSI amplitude calibration, CSI phase correction, and transformation from CSI to CIR as elaborated in
\ifappendixincluded
App.~\ref{app: data_preprocess})
\else
Appendix A)
\fi
to produce time‑domain CIR. This transformation is crucial for two reasons. First, these steps correct hardware‑induced distortions, reducing localization error (see Tab.~\ref{tab: result_RealData}). Secondly, it standardizes the input features across different datasets (e.g., WiFi dataset, high-precision real dataset, and synthetic dataset), ensuring consistency in the model's input characteristics.

The feature dimension of time-domain CIR data is defined as $F=b\times F_b$ where $b$ represents the number of APs and $F_b=2*N_{\mathrm{path}}$ corresponds to the number associated with the top $N_{\mathrm{path}}$ multipath components ranked by received power. Each selected path is characterized by its arrival time and received power, as recorded by each AP for a given fingerprint. 
The $N_{\mathrm{path}}$ determines the balance between capturing effective multipath components and including interference noise in the CIR representation. Consequently, this selection directly impacts the model’s training and prediction accuracy. The impact of $N_{\mathrm{path}}$ on model performance is analyzed in detail in Fig. \ref{fig: NumPaths_RMSE}.

\begin{table*}[ht]
\centering
\renewcommand{\arraystretch}{1.6}
\setlength{\tabcolsep}{4pt}
\begin{tabular}{|c|c|c|c|c|c|c|c|c|c|c|}
\hline
\multirow{2}{*}{\textbf{Scenario}} & \multirow{2}{*}{\textbf{Size}} & \multirow{2}{*}{\textbf{Heights}} & \multirow{2}{*}{\textbf{\# of APs}} & \multicolumn{3}{c|}{\textbf{Data Format}}                       & \multicolumn{3}{c|}{\textbf{Data Size}}                   \\ \cline{5-10}
                  &               &                  &                    & \textbf{WiFi} & \textbf{High-precision} & \textbf{Synthetic} & \textbf{WiFi} & \textbf{High-precision} & \textbf{Synthetic} \\ \hline
Hall              & 20m$\times$5m     & \multirow{2}{*}{\begin{tabular}[c]{@{}c@{}}Transmitters: 1.5m\\ Receivers: 2m\end{tabular}} & 5                  & \multirow{2}{*}{CSI to CIR}            & \multirow{2}{*}{CIR}                  & \multirow{2}{*}{CIR}                     & $N^0 = 500$          & $N^0 = 500$            & $m=12,N^i = 4000$            \\ \cline{1-2} \cline{4-4}\cline{8-10}
Lab               & 9m$\times$8m      &                  & 4                  &          &                   &                     & $N^0 = 135$          & Not Collected           & $m=10, N^i = 1200$            \\ \hline
\end{tabular}
\caption{Summary of Data Collection Setup}
\label{tab: Data Collection Setup}
\end{table*}

\subsubsection{Experimental Setup}

To evaluate the performance of the AGML model, we selected a range of well-established and widely used state-of-the-art models, including KNN~\cite{bahl2000radar}, WKNN~\cite{WKNN2020indoor}, CNN~\cite{ghozali2019indoor}, and MetaLoc~\cite{gao2022metaloc}. The detailed configurations of each model are summarized in 
\ifappendixincluded
App.~\ref{sec: model_configurations}.
\else
Appendix E.
\fi

\begin{table*}[t]
    \setlength\tabcolsep{6pt}
    \centering
    \renewcommand{\arraystretch}{1.4}
    \begin{tabular}{c|c|cccc|cccc}
    \hline
      \multicolumn{1}{c|}{\multirow{2}*{\textbf{Datasets}}} & \multicolumn{1}{c|}{\multirow{2}*{\textbf{Methods}}}    & \multicolumn{4}{c|}{\textbf{Scenario 1: Hall}} & \multicolumn{4}{c}{\textbf{Scenario 2: Lab}} \\
          \cline{3-10}
          & & $N_l=400$ & $N_l=100$ & $N_l=20$ & $N_l=5$ & $N_l=100$ & $N_l=50$ & $N_l=20$ & $N_l=5$ \\
         \hline
       \multicolumn{1}{c|}{\multirow{5}*{{$\{\ccalD^m\}$}}} & KNN~\cite{bahl2000radar} 
       & 0.6364 & 1.0853 & 3.1648 & 4.6829 & 1.1545 & 1.7839 & 3.9997 & 4.9962 \\
       & WKNN~\cite{WKNN2020indoor} 
       & 0.5682  & 0.9613 & 2.9394 & 4.3640 & 1.0093 & 1.4481 & 3.3759 & 4.7283 \\
       & CNN~\cite{ghozali2019indoor} 
       & 0.3689  & 0.5931 & 1.4482 & 3.3759 & 0.6043 & 0.9430 & 2.0812 & 3.5719 \\
       & MLP~\cite{gao2022metaloc} 
       & 0.3842 & 0.6319 & 1.5671 & 3.2946 & 0.5547 & 0.8834 & 1.9962 & 3.4698 \\
        & AGNN$_{N_l}$ (ours) 
       & 0.1968 & 0.4765 & 1.4912 & 3.1413 & 0.4046 & 0.7849 & 1.7384 & 3.3957 \\
       & \textbf{AGNN (ours)}
       & \textbf{0.1831}  & \textbf{0.3278} & \textbf{0.7391} & \textbf{2.5916} & \textbf{0.3812} & \textbf{0.5484} & \textbf{0.9366} & \textbf{2.9629} \\
         \hline
       \multicolumn{1}{c|}{\multirow{2}*{{$\{\ccalD^1, \dots, \ccalD^m\}$}}} & MetaLoc~\cite{gao2022metaloc} 
        & 0.3684 & 0.4247 & 0.5689 & 0.7219 & 0.5057 & 0.6845 & 0.8617 & 1.0728 \\
        & \textbf{AGML (ours)}  
        & \textbf{0.1794} & \textbf{0.2439} & \textbf{0.3478} & \textbf{0.4298} & \textbf{0.3386} & \textbf{0.4037} & \textbf{0.4846} & \textbf{0.5947} \\
    \hline
    \end{tabular}
    \caption{Averaged RMSE of all methods using synthetic data with various numbers of labeled fingerprints under two scenarios.}
    \label{tab: result_SynData}
\end{table*}

\subsection{Results on Synthetic Data}
\label{sec: result_SynData}

This section presents preliminary results on the synthetic datasets $\{\ccalD^1,\ccalD^2, \dots, \ccalD^m\}$ generated through computer simulations to mimic the original real-world scenario.

\vspace{4pt}
\textbf{\textit{- Superiority of AGNN.}}
First, we compare the localization performance of several methods without the meta-learning framework, using the $m$-th synthetic dataset, $\mathcal{D}^m$. Specifically, we sample $N^0$ fingerprints from $\mathcal{D}^m$ following Step-I in Sec.~\ref{sec: AGML_steps}, which includes $N_l$ labeled fingerprints and $N^0 - N_l$ unlabeled fingerprints, serving as the training data for the models. The remaining $N^m - N^0$ fingerprints are used as test data to assess the performance of the models.
The experimental results are presented in the upper part of Tab.~\ref{tab: result_SynData}, where AGNN$_{N_l}$ denotes the AGNN model trained using only the $N_l$ labeled fingerprints, similar to the supervised learning methods like KNN, WKNN, and vanilla MLP.

The results consistently demonstrate the superiority of AGNN over benchmark methods, showcasing the effectiveness of the GNN-based approach.  Specifically, when using only $N_l$ labeled fingerprints, AGNN$_{N_l}$ outperforms the benchmarks, especially when $N_l \geq 100$. This advantage arises from AGNN's capability to adaptively select suitable neighboring nodes via the ALM and to aggregate information through the learned adjacency matrix. As $N_l$ decreases, the physical distance between labeled fingerprints increases, causing the ALM to learn an adjacency matrix that approaches an identity matrix, which results in a significant increase in the localization error. In contrast, AGNN, which incorporates the $N^0 - N_l$ unlabeled fingerprints, shows superior performance in scenarios with very sparse labeled fingerprints. This improvement is not only owing to AGNN’s semi-supervised learning, which effectively incorporates information from unlabeled fingerprints, but also to the reduced physical distance between samples when these additional unlabeled data points are included, which enables the ALM to construct a more effective adjacency matrix, thereby enhancing information aggregation among neighboring nodes. Consequently, it demonstrates the significance of our proposed unlabeled fingerprint augmentation strategy. 

Moreover, AGNN demonstrates lower localization error in the Hall-scenario compared to the Lab-scenario, which is attributed to fewer obstacles (most fingerprints have LOS paths) and a higher density of fingerprints in the Hall-scenario, while the Lab-scenario has more obstacles (resulting in more NLOS paths) and a more sparse fingerprint distribution. Furthermore, AGNN exhibits minimal performance variation across the two scenarios, indicating the robustness of AGNN to NLOS noise.

\vspace{4pt}
\textbf{\textit{Validity of Meta-learning Framework.}}
We further evaluate the performance of meta-learning-based methods, MetaLoc and AGML, across all synthetic datasets, $\{\mathcal{D}^1, \mathcal{D}^2, \dots, \mathcal{D}^{m}\}$. Specifically, MetaLoc and AGML both require meta-training using the synthetic datasets $\{\mathcal{D}^1, \mathcal{D}^2, \dots, \mathcal{D}^{m-1}\}$, followed by meta-testing on the same $N^0$ fingerprints sampled from $\mathcal{D}^m$. The experimental results are presented in the lower part of Tab.~\ref{tab: result_SynData}. 
Two key observations can be drawn from these results. 1) Both MetaLoc and AGML outperform non-meta-learning methods like MLP and AGNN, achieving lower RMSE across all $N_l$ values, particularly when $N_l \leq 20$. This is because, while a larger number of labeled fingerprints sufficiently trains the models, as $N_l$ decreases, meta-learning leverages data from similar scenarios to address data scarcity and reduce RMSE. 2) AGML consistently achieves the lowest localization error, with a smaller increase in error as $N_l$ decreases compared to other methods, indicating its high effectiveness and reduced dependency on labeled fingerprints for accurate localization.

\begin{figure}[t] 
\centering
\includegraphics[width=0.9\linewidth]{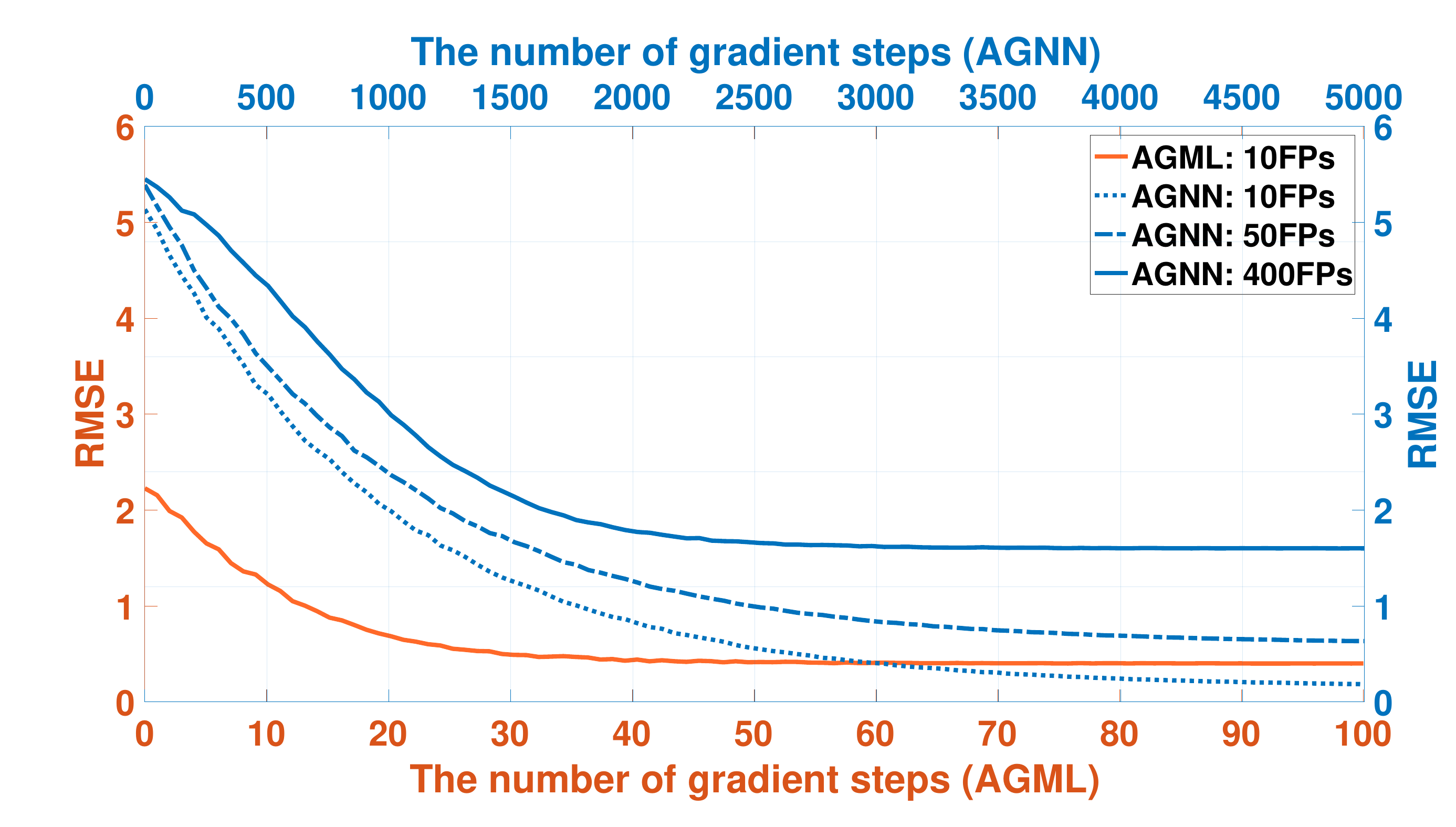}
\caption{Comparison of convergence between AGML and AGNN in terms of localization error (RMSE).}
\label{fig: AGML_Convergence_Sim}
\end{figure}

\vspace{4pt}
\textbf{\textit{Convergence Speed of AGML.}}
To evaluate the convergence speed of AGML, we compare its performance to AGNN in terms of localization error (RMSE), as shown in Fig.~\ref{fig: AGML_Convergence_Sim}. Specifically, AGML performs meta-training using simulated datasets $\{\ccalD^1, \dots, \ccalD^{m-1}\}$, followed by model adaptation using 10 fingerprints from $\ccalD^m$. The RMSE of the AGML during the meta-test is represented by the orange curve, while AGNN's performance is shown with blue curves for 10, 50, and 400 fingerprints, respectively.

Several key insights emerge from this comparison: Firstly, AGNN's RMSE decreases as the number of labeled fingerprints increases, achieving lower error rates with more data. However, this improvement is obtained at the cost of a slower convergence rate. For instance, AGNN with 400 fingerprints requires significantly more gradient steps to stabilize, demonstrating the trade-off between data quantity and training time.
Conversely, AGML exhibits a more significant reduction in RMSE during the initial gradient steps, indicating rapid convergence to lower localization error. This quick adaptation results from the knowledge gained during meta-training, enabling AGML to efficiently adjust its model parameters using only a small number of fingerprints from $\ccalD^m$. This capability manifests AGML's strong generalization across datasets, even with very limited labeled data.
Notably, AGML reaches a similar RMSE after around 40 gradient steps as AGNN operates with 400 fingerprints after over 3000 gradient steps. This demonstrates AGML's effectiveness in adapting quickly and achieving low error with minimal data, making it highly suitable for environments where labeled data is scarce.

\subsection{Results on Real-world Data}
\label{sec: result_RealData}

This section presents the results on the real-world datasets $\{\ccalD^0,\ccalD^t\}$, which consist of fingerprints collected in two indoor environments (Hall-scenario and Lab-scenario) using two types of devices: consumer‑grade WiFi devices and high-precision professional instruments. 

The experimental results are summarized in Tab.~\ref{tab: result_RealData}. Here, AGNN-DP denotes the AGNN model using the raw CSI data as input without the data preprocessing discussed in 
\ifappendixincluded
App.~\ref{app: data_preprocess},
\else
Appendix A,
\fi
in contrast to the other methods that rely on CIR data derived from processed CSI.
In the sequel, AGML-DA refers to AGML meta-trained on synthetic data and directly meta-tested on real-world data without accounting for distribution alignment.
Additionally, AGML$_{\ccalD^m}$ represents AGML meta-trained only on the $m$-th synthetic dataset $\ccalD^m$ without considering the synthetic environmental perturbations.

\begin{table*}[t]
    \setlength\tabcolsep{2pt}
    \centering
    \renewcommand{\arraystretch}{1.4}
    \begin{tabular}{c|c|ccc|ccc|ccccc}
    \hline
      \multicolumn{1}{c|}{\multirow{2}*{\textbf{Datasets}}} & \multicolumn{1}{c|}{\multirow{2}*{\textbf{Methods}}}    & \multicolumn{3}{c|}{\textbf{WiFi (Hall)}} & \multicolumn{3}{c|}{\textbf{WiFi (Lab)}} & \multicolumn{5}{c}{\textbf{High-precision (Hall)}} \\
          \cline{3-13}
          & & $N_l=400$ & $N_l=100$ & $N_l=20$  & $N_l=100$ & $N_l=50$ & $N_l=20$ & $N_l=400$ & $N_l=100$ & $N_l=50$ & $N_l=20$ & $N_l=5$ \\
         \hline
       \multicolumn{1}{c|}{\multirow{6}*{{$\{\ccalD^0,\ccalD^t\}$}}} 
       & KNN~\cite{bahl2000radar} 
       & 1.0719 & 1.9292 & 4.9791 & 2.4910 & 3.5719 & 5.1712
       & 0.8339 & 1.5190 & 2.7819 & 3.9717 & 5.6924\\
       & WKNN~\cite{WKNN2020indoor} 
       & 0.8391  & 1.4385 & 3.9619 & 2.0791 & 3.2841 & 4.3872
       & 0.7104 & 1.2829 & 2.3948 & 3.6191 & 5.4919\\
       & CNN~\cite{ghozali2019indoor} 
       & 0.5583  & 0.9138 & 3.0075 & 1.4684 & 2.5890 & 3.8836
       & 0.4102 & 0.7139 & 1.2940 & 2.1873 & 3.3391\\
       & MLP~\cite{gao2022metaloc} 
       & 0.5909 & 0.9510 & 3.1819 & 1.3381 & 2.3532 & 3.6291 
       & 0.4691 & 0.7892 & 1.4829 & 2.3791 & 3.5910\\
       & \textbf{AGNN (ours)} 
       & \textbf{0.4524}  & \textbf{0.7468} & \textbf{2.3840} & \textbf{0.8381} & \textbf{1.7417} & \textbf{2.9912} & \textbf{0.2638} & \textbf{0.3981} & \textbf{0.7461} & \textbf{1.6629} & \textbf{3.2840}  \\
       & AGNN-DP (ours)
       & 0.6183 & 1.0489 & 3.414 & 1.1738 & 1.9972 & 3.7217 & N.A. & N.A. & N.A. & N.A. & N.A. \\
         \hline
       \multicolumn{1}{c|}{\multirow{4}{*}{$\begin{array}{c}\{\ccalD^1, \dots, \ccalD^m\} \\ 
       \{\ccalD^0,\ccalD^t\}\end{array}$}}
        & MetaLoc~\cite{gao2022metaloc} 
        & 0.5425 & 0.7739 & 1.5822 & 0.9391 & 1.7674 & 2.5681
        & 0.4396 & 0.5048 & 0.7839 & 1.0472 & 2.6239 \\
        & AGML-DA (ours)
        & 0.4292 & 0.5899 & 0.9747 & 0.7782 & 1.3791 & 2.0291
        & 0.2482 & 0.3498 & 0.6404 & 0.9846 & 1.7749 \\
         & AGML$_{\ccalD^m}$ (ours)
        & 0.3902 & 0.5229 & 0.8102 & 0.6832 & 1.2138 & 1.9321
        & 0.2363 & 0.2927 & 0.4189 & 0.5743 & 1.1028 \\
        & \textbf{AGML (ours)}  
        & \textbf{0.3824} & \textbf{0.5178} & \textbf{0.7782} & \textbf{0.6791} & \textbf{1.1749} & \textbf{1.8826} 
        & \textbf{0.2335} & \textbf{0.2892} & \textbf{0.4069} & \textbf{0.5429} & \textbf{1.0361} \\
    \hline
    \end{tabular}
    \caption{The averaged loss (RMSE) of all methods using real-world data with various numbers of labeled fingerprints under different scenarios.}
    \label{tab: result_RealData}
\end{table*}

\begin{figure}[t] 
\centering
\includegraphics[width=0.9\linewidth]{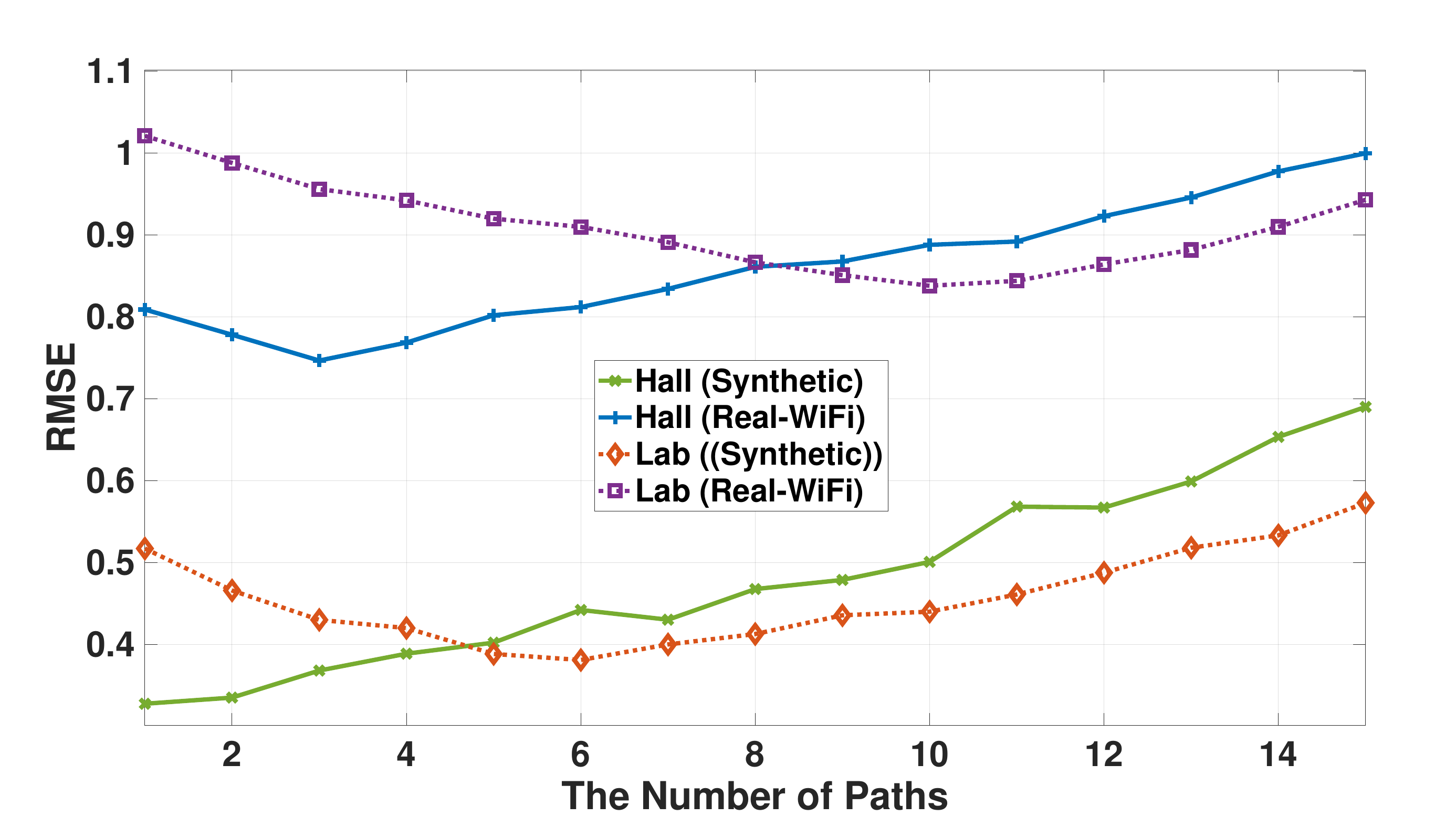}
\caption{RMSE versus the number of paths $N_{\mathrm{path}}$ under different scenarios for AGNN with $N_l=100$ samples.}
\label{fig: NumPaths_RMSE}
\end{figure}

\vspace{4pt}
\textbf{\textit{Performance Analysis for WiFi Data.} }
The evaluation of model performance on real-world WiFi data yields several notable observations. As shown in the left part of Tab.~\ref{tab: result_RealData}, the AGNN model consistently outperforms traditional models (KNN, WKNN, MLP), with the performance gap enlarging as the number of labeled fingerprints, $N_l$, decreases. This advantage mirrors the trends observed in the synthetic data analysis, where AGNN effectively leveraged spatial information through its learned adjacency matrix obtained using our proposed unlabeled fingerprint augmentation strategy. However, it is important to note that AGNN's performance on WiFi data is inferior to its performance on synthetic data. This discrepancy can be attributed to the lower quality of WiFi signals, which tend to be noisier and more susceptible to NLOS propagation conditions. The increased noise level and signal degradation hinder the model's ability to accurately construct adjacency matrices, ultimately affecting its localization accuracy.
The AGNN-DP model, which directly utilizes raw CSI data without preprocessing, exhibits lower performance than the AGNN model, which uses CIR data transformed from the preprocessed CSI. This result demonstrates the critical role of data preprocessing in extracting meaningful features from raw CSI, which often contains redundant or noisy components that can degrade localization performance when left unprocessed.

In contrast to MLP and AGNN, the meta-learning models (MetaLoc and AGML) demonstrate significant performance improvements, particularly in scenarios where labeled fingerprints are scarce. This finding suggests that the meta-learning frameworks can effectively transfer knowledge acquired from synthetic training scenarios to real-world WiFi localization tasks. Such knowledge transfer helps to mitigate the challenges posed by the lower quality and very sparse labeled data in real-world settings.

\begin{figure}[t]
    \centering
    \begin{subfigure}[b]{0.9\linewidth}
        \centering
        \includegraphics[width=\linewidth]{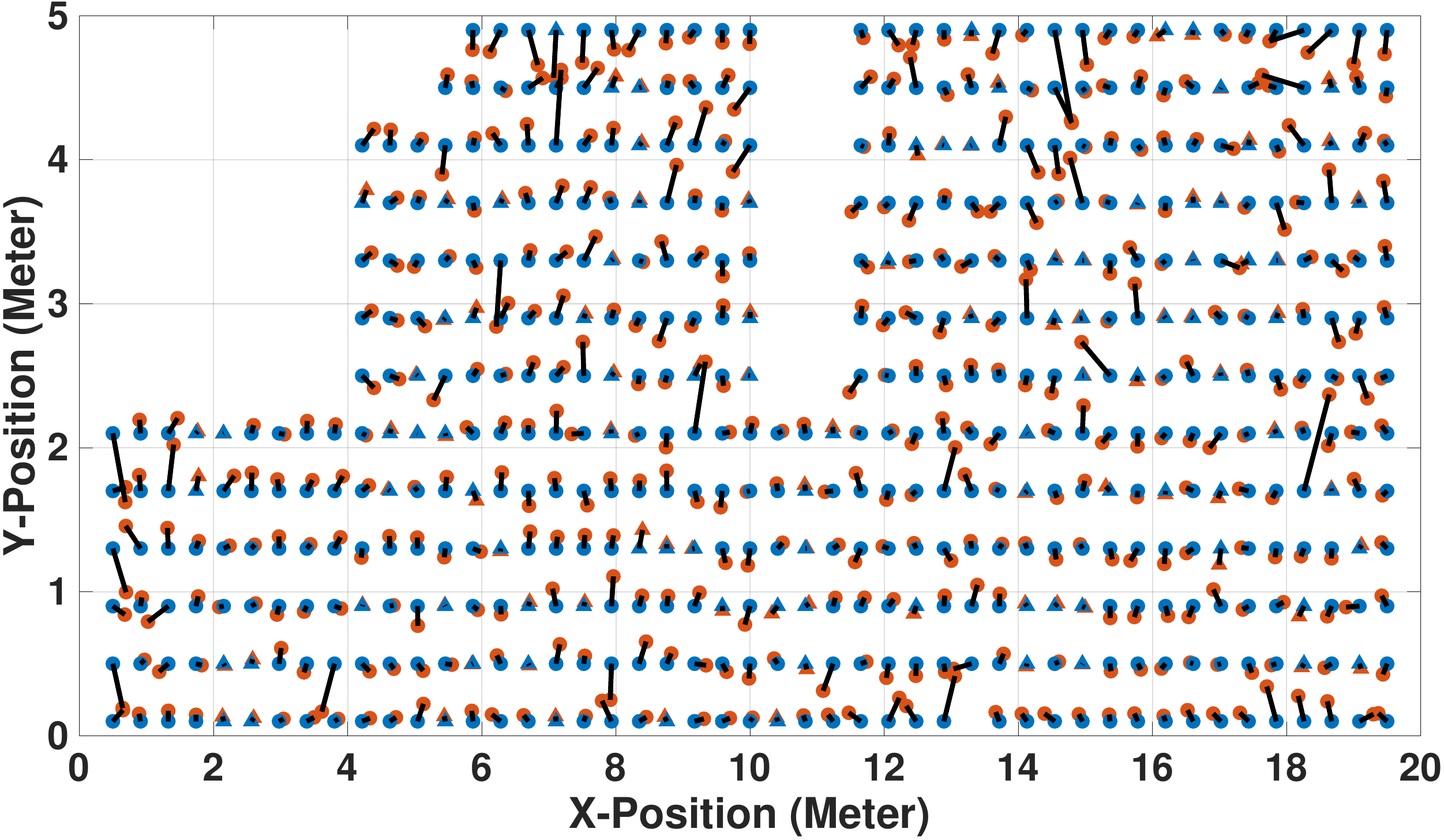}
        \caption{Scatter plot of predicted and true positions. Herein, blue nodes represent the true positions, while orange nodes indicate the predicted positions. Triangles denote fingerprints, whereas circles represent TPs. Additionally, black lines depict localization errors between the true positions and those predicted by AGML.}
        \label{fig: scatter_plot}
    \end{subfigure}

    \vspace{0.5cm} 

    \begin{subfigure}[b]{0.9\linewidth}
        \centering
        \includegraphics[width=\linewidth]{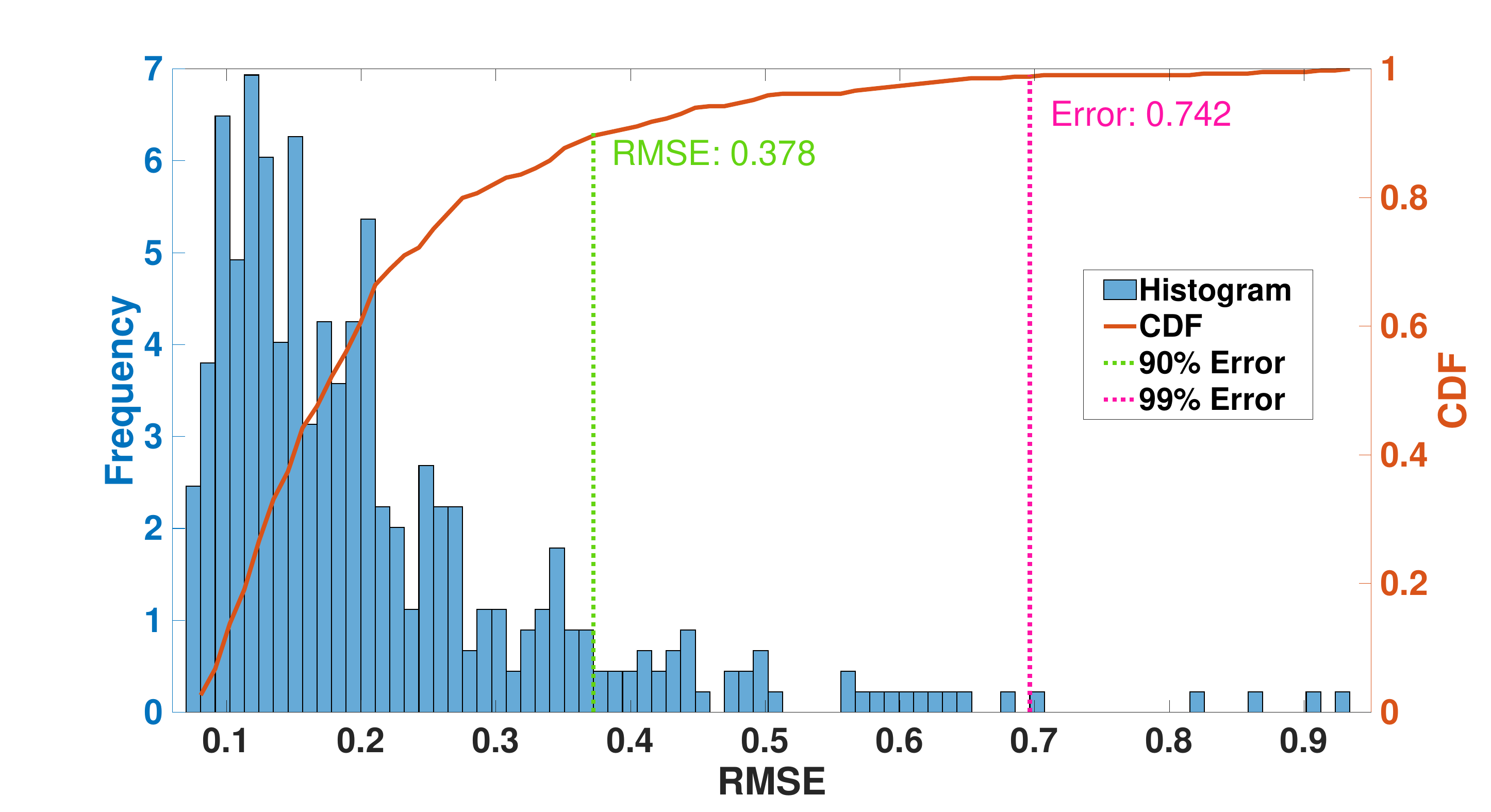}
        \caption{Histogram and CDF of the TPs' RMSE.}
        \label{fig: histogram_cdf}
    \end{subfigure}
    
    \caption{Visualization of AGML localization performance in the Hall-scenario using the high-precision real dataset with $N_l = 100$.}
    \label{fig: combined_figure}
\end{figure}

\vspace{4pt}
\textbf{\textit{Effect of $N_{\mathrm{path}}$.} }
We conducted experiments to investigate the relationship between the number of multipaths, $N_{\mathrm{path}}$, in the input CIR and the localization error of AGNN across different datasets, as shown in Fig.~\ref{fig: NumPaths_RMSE}. The experimental results reveal that the optimal value of $N_{\mathrm{path}}$ corresponding to the lowest RMSE achieved by the AGNN model varies across different datasets. Specifically, in the synthetic Hall-scenario, where LOS transmission dominates, the received power and arrival time from the main path, corresponding to the maximum received power, are sufficient to provide adequate distance information. Therefore, when $N_{\mathrm{path}}=1$, the AGNN model achieves the best localization performance. As $N_{\mathrm{path}}$ increases, additional paths provide redundant features, which actually degrade the AGNN’s localization accuracy. In contrast, in the other three scenarios, the localization error of the AGNN model initially decreases and then increases as $N_{\mathrm{path}}$ increases. This behavior is primarily due to the fact that paths with lower received power typically carry more NLOS noise. Thus, while introducing new paths increases the number of effective features, it also introduces additional noise, leading to a trade-off between multipath feature information and noise.
Furthermore, in both real WiFi and synthetic data, the optimal $N_{\mathrm{path}}$ for achieving the lowest RMSE in the Hall-scenario is smaller than in the Lab-scenario. This difference arises because NLOS dominates in the Lab-scenario, where even the path with the maximum received power in most fingerprint points remains an NLOS path. Consequently, incorporating more multipath information is necessary to improve localization accuracy for AGNN.

\vspace{4pt}
\textbf{\textit{Performance Analysis for High-precision Real Data.} }
The analysis of model performance on high-precision real data, as presented in the right part of Tab.~\ref{tab: result_RealData}, further emphasizes the strengths of meta-learning approaches. 
Here, the results for AGNN-DP are marked as not available (N.A.) because the raw high-precision real data is collected in CIR format, eliminating the need for preprocessing. 
Both AGNN and AGML achieve significantly lower RMSE values compared to the WiFi dataset, reflecting the superior data quality provided by high-precision professional measurements. 
For visualization, the predicted positions obtained from the AGML are depicted in Fig.~\ref{fig: combined_figure} (a) along with the true positions. Additionally, the histogram and Cumulative Distribution Function (CDF) of TPs' RMSE for the corresponding scenario are provided in Fig.~\ref{fig: combined_figure} (b).
The higher fidelity and finer granularity of the data enable these models to construct more precise adjacency matrices, enabling more accurate spatial relationships for localization. 

Among the evaluated methods, AGML consistently achieves the lowest localization errors across all scenarios, particularly in cases where labeled fingerprints are limited (e.g., $N_l=5$). This exceptional performance reinforces the effectiveness of the AGML framework. Furthermore, the observed performance gap between AGML and AGML-DA highlights that AGML not only benefits from the knowledge acquired during the meta-training phase but also effectively adapts to the specific distributional characteristics encountered during meta-testing. This adaptability is particularly pronounced when working with high-quality datasets.

\begin{figure}[t] 
\centering
\includegraphics[width=1\linewidth]{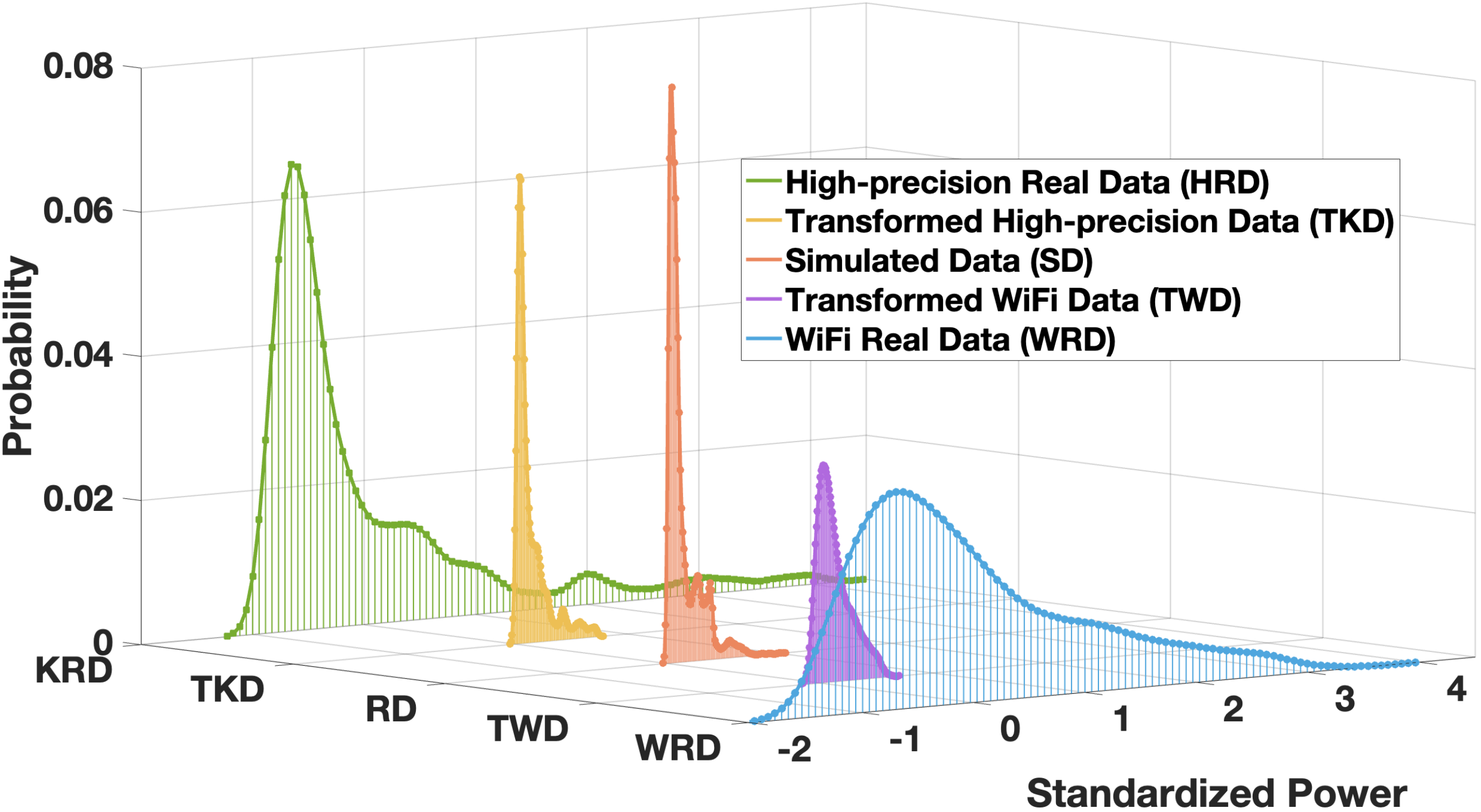}
\caption{Comparison of distributions before and after distribution alignment using high-precision real data and WiFi data.}
\label{fig: Distribution_Shift}
\end{figure}

\vspace{4pt}
\textbf{\textit{Significance of Distribution Alignment.} }
To validate the impact of distribution alignment between synthetic and real-world datasets on model performance, we use the strongest power received by the $4$-th AP as an example, plotting the distributions of the raw WiFi data and high-precision real data, their transformed distributions after distribution alignment, and the distribution of synthetic data, as shown in Fig.~\ref{fig: Distribution_Shift}. In this figure, we observe that the distributions of the synthetic and high-precision real data are similarly shaped, while the WiFi data deviates significantly. After applying the distribution alignment, the transformed distribution of high-precision real data closely aligns with the distribution of synthetic data, whereas the WiFi data still shows considerable differences. This explains AGML's superior performance on high-precision real data, where it achieves low RMSE even with sparse labeled fingerprints, and its degraded performance on WiFi data, where the mismatched distribution introduces feature distortions that hinder model effectiveness.
Moreover, the results from the high-precision real dataset reveal that when the quality of real-world data is high and its distribution closely matches that of the synthetic data, the performance gap between AGML and AGML-DA diminishes. This indicates that exploiting synthetic data to better reflect real-world conditions can effectively reduce the impact of distribution mismatch, thereby enhancing the generalization of meta-learning models.

\vspace{4pt}
\textbf{\textit{Impact of Perturbations on Synthetic Dataset.}} 
Finally,  we investigate the impact of perturbations introduced in synthetic labeled fingerprint augmentation on the localization accuracy of the AGML model. 
As shown in Tab.~\ref{tab: result_RealData}, AGML$_{\ccalD^m}$ and AGML correspond to models meta-trained using only the $m$-th synthetic dataset $\ccalD^m$ and all synthetic datasets $\{\ccalD^1,\dots, \ccalD^m\}$, respectively. The results indicate that AGML consistently outperforms AGML$_{\ccalD^m}$ across all scenarios.

\begin{figure}[t] 
\centering
\includegraphics[width=1\linewidth]{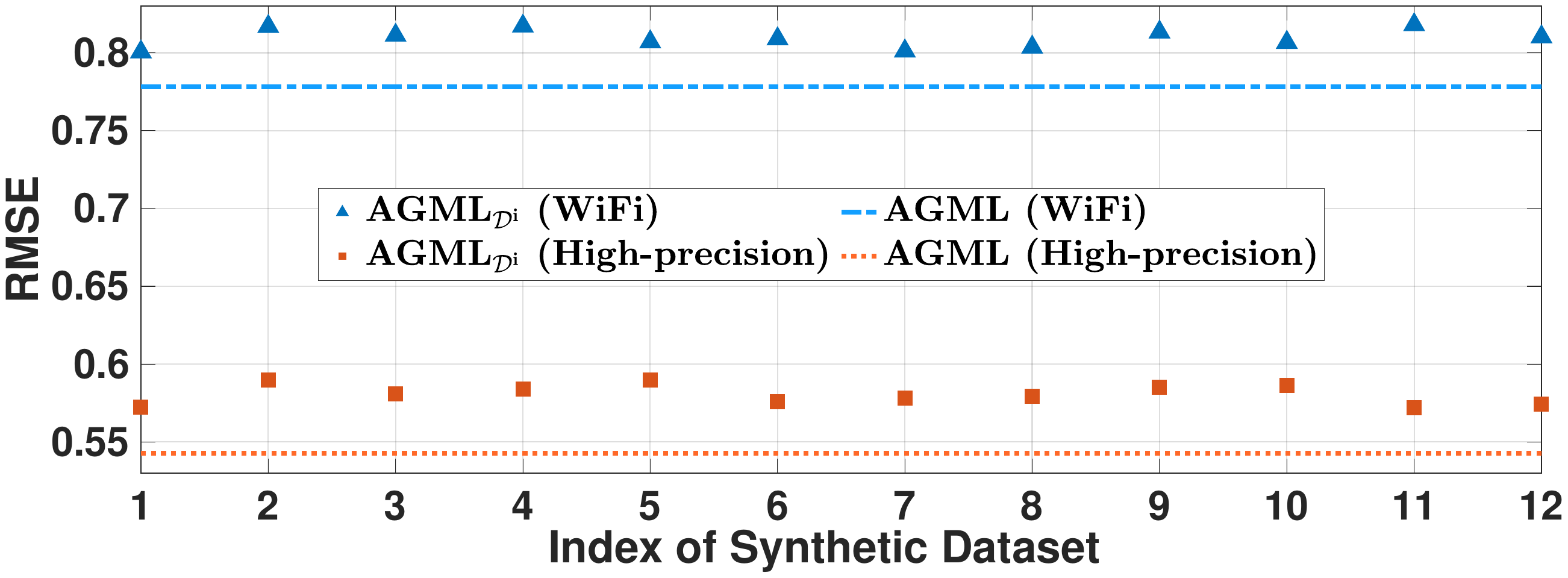}
\caption{RMSE results between AGML and AGML$_{\ccalD^i}, \forall i \in [m]$, in the Hall-Scenario using both WiFi and high-precision real datasets.}
\label{fig: Perturbation_SynData}
\end{figure}

To further validate this observation, we conduct additional experiments in the Hall-Scenario using both WiFi and high-precision real datasets. Specifically, we compare the RMSE of AGML trained on a single synthetic dataset, AGML$_{\ccalD^i}$, with that of AGML trained on all synthetic datasets, as presented in Fig.~\ref{fig: Perturbation_SynData}.
In this setup, $\ccalD^1$ is a synthetic dataset generated by strictly following the environmental characteristics of the Hall-Scenario, while the remaining synthetic datasets are generated by introducing slight perturbations to the original setting. Experimental results demonstrate that AGML consistently outperforms AGML$_{\ccalD^i}, \forall i \in [m]$, even when AGML$_{\ccalD^1}$ is trained on a dataset that precisely matches the environmental characteristics of the Hall-Scenario.
This performance gap between AGML and AGML$_{\ccalD^i}$ stems from the inherent limitations of synthetic environments. Specifically, simulated scenarios cannot fully capture the complexity and variability of real-world environmental characteristics, such as obstacle placement and transmitter/receiver orientations. Consequently, meta-training on a single synthetic dataset limits the model's generalization. In contrast, meta-training on multiple synthetic datasets enables the model to capture variations of environmental characteristics among different datasets. This strategy enhances the robustness of AGML during meta-learning, leading to improved generalization and localization accuracy in real-world environments.

\section{Conclusion}
\label{sec: conclusion}

Fingerprint‑based indoor localization with extremely sparse fingerprints remains challenging due to laborious data collection and complex signal propagation. This paper has introduced the AGML model to address these challenges. First, we formalized the localization scenario and proposed two complementary data augmentation strategies: 1) unlabeled fingerprint augmentation via mobile platforms, enabling semi‑supervised learning on abundant unlabeled data, and 2) synthetic labeled fingerprint augmentation through digital twins, together with a distribution alignment to reduce the feature discrepancy between synthetic and real-world measurements. Next, we tailored an AGNN to autonomously learn graph structure and node aggregation weights from a dataset with limited labeled and abundant unlabeled fingerprints. Building on this, we embedded AGNN within a meta-learning framework to enable rapid adaptation from synthetic datasets with slight perturbations to the target real-world scenario. Extensive experiments on both synthetic and real‑world datasets (consumer‑grade Wi‑Fi and high‑precision professional equipment) in LOS and NLOS settings demonstrate that AGML consistently outperforms state‑of‑the‑art baselines, achieving substantially lower RMSE, faster convergence, and robust performance with very few labeled points.

\bibliographystyle{IEEEtran}
\bibliography{Refs.bib}

\vfill

\ifappendixincluded
\appendix
\section{Appendix}
\subsection{Data Preprocessing for WiFi Platform}
\label{app: data_preprocess}

\subsubsection{Transformation Between CSI and CIR}
Both frequency-domain CSI and time-domain CIR  are essential for capturing the small-scale multipath effects in wireless channel measurements. Specifically, each CSI is described as 
\begin{equation}
H\left(f_k\right)=\left\|H\left(f_k\right)\right\| e^{j \sin (\angle H(f_k))}, k=1, \ldots, K, \label{eq:csi}
\end{equation}
where $\left\|H\left(f_k\right)\right\|$ denotes the amplitude and $\angle H(f_k)$ denotes the phase of the CSI at the central subcarrier frequency $f_k$. The total number of subcarriers $K$ depends on the channel bandwidth specified by IEEE 802.11ac standards: 64 for 20 MHz, 128 for 40 MHz, and 256 for 80 MHz. Meanwhile, CIR $h(\varsigma)$ can be defined as
\begin{equation}
h(\varsigma)=\sum_{i=1}^N a_i e^{-j \theta_i} \delta\left(\varsigma-\varsigma_i\right), \label{eq:cir}
\end{equation}
where the $i$-th term in the CIR corresponds to a propagation path with a time delay $\varsigma_i$, amplitude attenuation $a_i$, and phase shift $\theta_i$, with $N$ denoting the total number of paths. In brief, CSI and CIR provide distinct perspectives on channel characteristics. Frequency-domain CSI can be transformed into time-domain CIR using Inverse Fast Fourier Transform (IFFT)~\cite{8423070,yang2013rssi}. While many studies have primarily focused on either the amplitude~\cite{CRISLoc} or phase~\cite{zhu2022intelligent} of CSI, often leading to the loss of valuable information. Instead, we consider both the CSI amplitude and phase simultaneously. This combination allows us to convert the full CSI information into CIR via IFFT, preserving more detailed channel characteristics.

\subsubsection{Abnormal Packet Removal}
Fig.~\ref{fig:csi-example} displays CSI measurements obtained from the Asus RT-AC86U WiFi device at a stationary point under 80MHz bandwidth. Specifically, Fig.~\ref{fig:csi-example} (a) and (b) show the amplitude and phase of CSI across subcarriers for the first packet, respectively. Fig.~\ref{fig:csi-example} (c) illustrates a CSI amplitude heatmap for all 900 collected packets, where lighter areas indicates higher amplitudes. It is noted that guard subcarriers from -128 to -123, -1 to 1, and 123 to 127 may contain random values, so they are typically set to zero to prevent disturbance as recommended in~\cite{nexmon:project}. 
\begin{figure}
    \centering
    \includegraphics[scale=0.55]{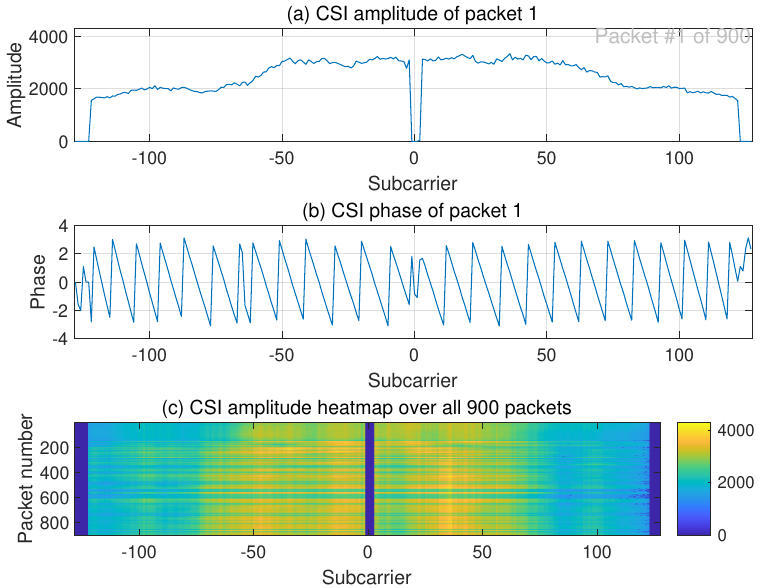}
    \caption{CSI measurements collected from the commercial WiFi device Asus RT-AC86U in a stationary point: (a) CSI amplitude across subcarriers for the first packet.
(b) CSI phase across subcarriers for the first packet. (c) CSI amplitude heatmap over all 900 collected packets, with lighter areas indicating higher amplitudes.
}
\label{fig:csi-example}
\end{figure}
As shown in Fig.~\ref{fig:csi-example} (c), there are some abnormal packets due to environmental noise. To mitigate this issue, we adopt an abnormal packet removal approach utilizing the Mahalanobis distance as shown in
\begin{equation}
d(x) = (x - \mu)^T \Sigma^{-1} (x - \mu),
\end{equation}
where $x \in \mathbb{R}^{K}$ is a column vector representing a CSI packet sample, $ \mu \in \mathbb{R}^{K} $ is the mean vector of the CSI observations, and $ \Sigma \in \mathbb{R}^{ K \times K} $ is the covariance matrix of the subcarrier measurements. The Mahalanobis distance quantifies the deviation of a CSI packet $x$ from a given distribution, incorporating the covariance among vector elements. A CSI packet with a higher Mahalanobis distance indicates an outlier. By eliminating the top 10\% of packets with the largest Mahalanobis distances, we ensure that the remaining CSI packets are suitable for accurate fingerprinting.

\subsubsection{CSI Amplitude Calibration}
\begin{figure}
	\centering
	\subfloat[Raw CSI amplitude heatmap for AP1]{\includegraphics[scale=0.26]{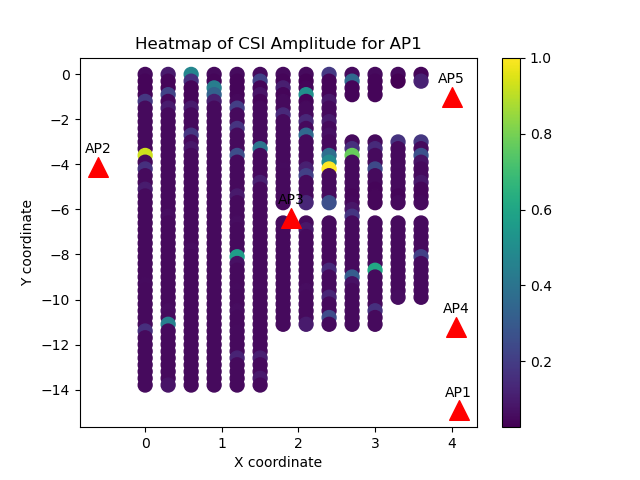}}
	\hfill 
	\subfloat[Calibrated CSI amplitude heatmap for AP1]{\includegraphics[scale=0.26]{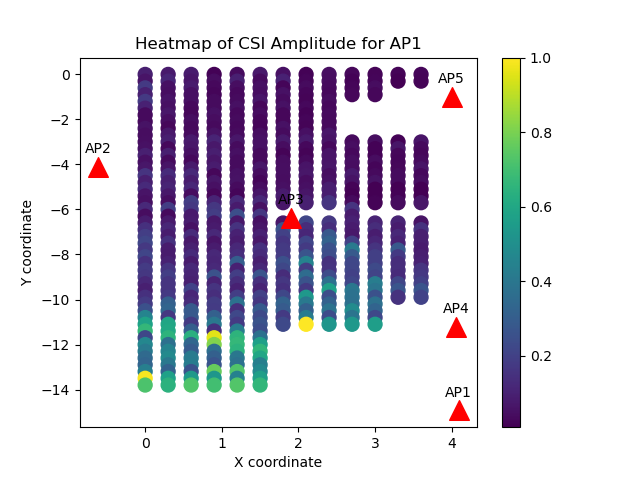}}
	\caption{AP1 located in the lower right corner of the picture.}
	\label{fig:amplitude}
\end{figure}

The left subplot of Fig.~\ref{fig:amplitude} shows a heatmap of normalized raw CSI amplitudes transmitted from a mobile device and received by AP1, which is located in the bottom right corner of the area. In contradiction to the basic path-loss principle, areas closer to AP1 exhibit lower amplitudes, while some points far from AP1 unexpectedly show higher amplitude values. This occurs because the Automatic Gain Control (AGC)~\cite{CRISLoc} of receiver modifies CSI amplitude, with the adjustment factor $s$ changing according to channel conditions. To address this, we calibrate raw CSI amplitude using RSS since it is captured before AGC adjustment. Notably, on our adopted Nexmon CSI platform~\cite{gringoli2019free}, each CSI packet header includes a byte for RSSI, eliminating the need for additional RSSI collection equipment. 

As a Linear Time-Invariant (LTI) system, AGC uniformly affects all subcarriers. Based on this principle, we apply the same adjustment coefficient 
$s$ to all subcarriers of CSI, as proposed in~\cite{CRISLoc}
\begin{equation}
s=\sqrt{\frac{10^{R S SI / 10}}{\sum \left\|H\left(f_k\right)\right\|^2}},
\end{equation}
where $\left\|H\left(f_k\right)\right\|$ represents the amplitude of the 
$k$-th subcarrier extracted from CSI, and RSSI is the received signal strength in dB. By employing this method, raw CSI amplitudes are rescaled to match the corresponding RSSI, effectively canceling the effects of AGC. The right subplot of Fig.~\ref{fig:amplitude} shows a heatmap of normalized calibrated CSI amplitudes, we can see that the calibrated amplitudes increase closer to AP1 and decrease with distance, consistent with the basic path-loss principle. At each location, we select 50\% of the CSI packets after filtering out abnormal ones to calculate the adjustment coefficient $s$, and the rest packets are adjusted by $s$ for localization purposes.
\subsubsection{CSI Phase Correction}
Fig.~\ref{fig:csi-example}
 (b) displays the raw CSI phase for the first packet, spanning the interval $(-\pi,\pi)$. We can see that when the phase exceeds this interval, it results in phase jumps. To address this, we unwrap the raw phase to extend its range from $-\infty$ to $\infty$.  The formula for the unwrapped CSI phase is given by~\cite{8423070}:
\begin{equation}
\phi_k = 2\pi(f_0 + k\Delta f)\frac{d}{c} = (2\pi\Delta f\frac{d}{c})k + 2\pi f_0\frac{d}{c},
\end{equation}
where $d$ represents the path distance, $c$ is the speed of light, $k$ denotes the subcarrier index, and $\Delta f$ is the subcarrier spacing, set at 312.5 KHz for 802.11ac standards.  Utilizing this model, we apply least squares linear regression for estimation. The left subplot in Fig.~\ref{fig:phase-fitting} illustrates the averaged unwrapped phase of all packets, shown with a blue line, while the linear regression fitting result is shown in red. The right subplot compares the original wrapped phases with their linear fits.
\begin{figure}
    \centering
    \includegraphics[scale=0.45]{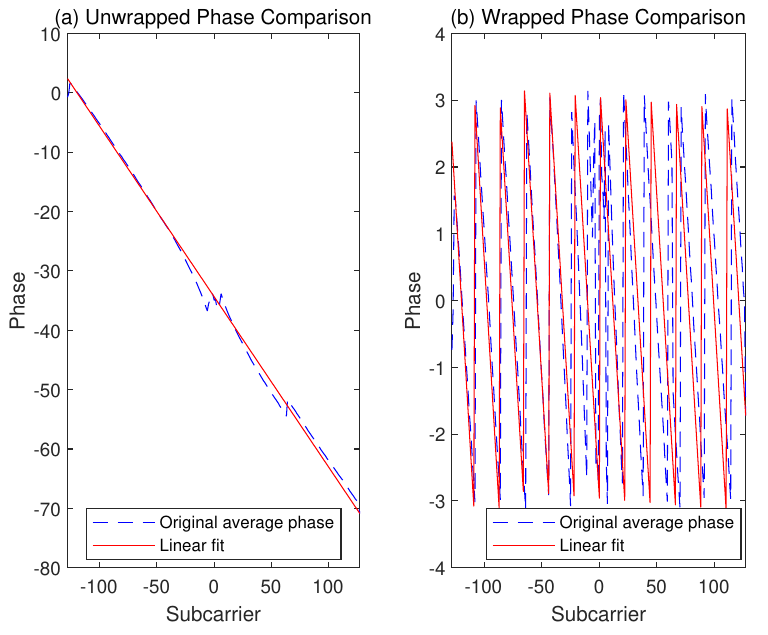}
    \caption{CSI phase processing: (a) unwarpped phase (b) wrapped phase. }
    \label{fig:phase-fitting}
\end{figure}

\begin{figure}
    \centering
    \includegraphics[scale=0.32]{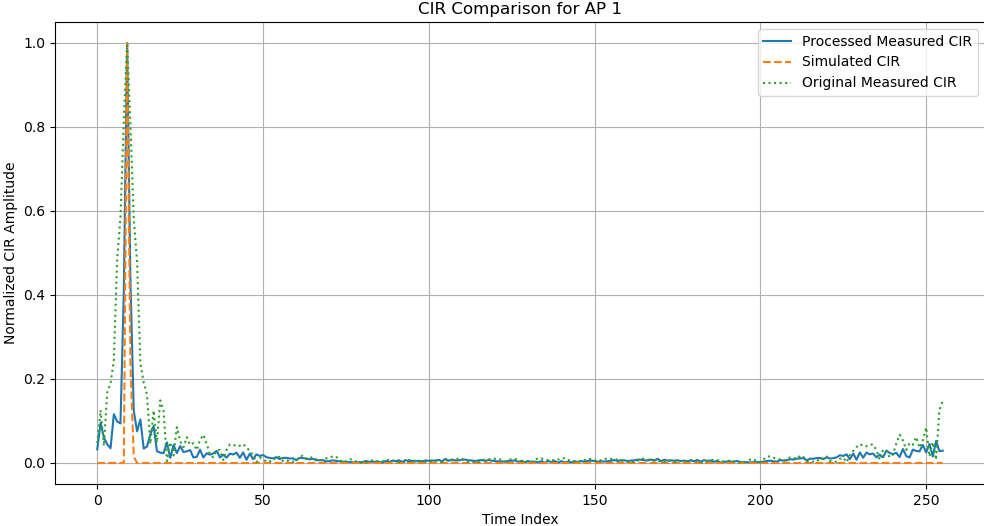}
    \caption{CIR Comparison among transformed by original CSI IFFT (original measured CIR marked as green line), processed CSI IFFT (processed measured CIR marked with blue line), and simulated CIR (marked with orange line) generated by the WI platform at the same location.}
    \label{fig:CIR-comparison}
\end{figure}
Following the CSI IFFT transformation, we derive the CIR as a series of signal samples in the time domain, as illustrated in \cite{8423070}, the time resolution $\Delta \varsigma$ of the CIR is inversely proportional to the bandwidth $B$ of the CSI as $\Delta \varsigma \propto \frac{1}{B}$. This relationship implies that a wider bandwidth results in a higher resolution for the CIR.

Fig.~\ref{fig:CIR-comparison} provides a comparison of the CIR obtained from both the original and processed CSI. The original measured CIR is marked by a green line, while the processed measured CIR is represented by a blue line. Additionally, the simulated CIR generated by the WI platform at the same location for a fixed AP1 is depicted with an orange line. The comparison reveals that the CIR derived from unprocessed CSI through IFFT exhibits some noise peaks, particularly at the end of the time index. In contrast, the CIR derived from processed CSI presents a cleaner profile, closely resembling the simulated CIR.

\subsection{Setups of WiFi Communication Platform}
\label{app: uplink-communication-framework}

For the WiFi setup, we leverage the Nexmon CSI platform on commercial WiFi devices, which provides an accessible and cost-effective means of capturing the frequency-domain CSI from 802.11ac WiFi frames. 
We use a Nexus 5 smartphone as the transmitting device and five Asus RT-AC86U routers as APs. All routers are connected to a central hub, forming an uplink communication system, as shown in Fig.~\ref{fig:uplink-communication-framework}. This system operates in the 5 GHz band, using channel 157 (at 5785 MHz) with an 80 MHz bandwidth.

\begin{figure}[t]
    \centering
    \includegraphics[scale=0.45]{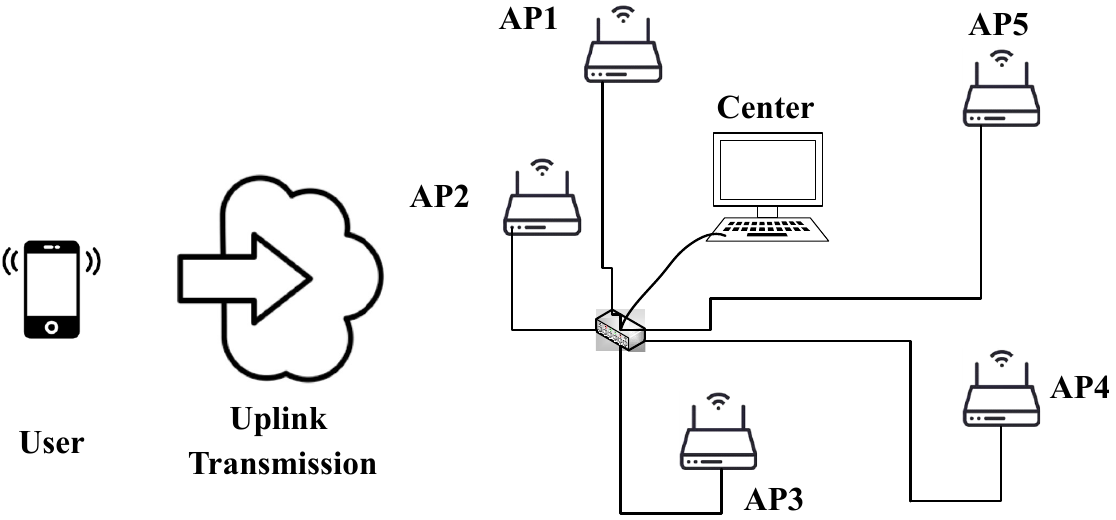}
    \caption{The overall uplink communication framework consisting of five APs as receivers and a smartphone as the transmitter.}
    \label{fig:uplink-communication-framework}
\end{figure}

\subsection{Setups of Professional Communication Platform}
\label{app: setups-keysight}
To further validate our method, we utilize high-precision Keysight wireless channel measurement equipment.
The Keysight N5182B signal generator and N9030B spectrum analyzer enable the collection of more accurate channel information, particularly under high-precision conditions. Fig.~\ref{fig:keysight-cir} shows CIR measurements collected using the Keysight platform under 40 MHz and 100 MHz bandwidths. The 100 MHz bandwidth offers higher time resolution, which is beneficial for capturing detailed multipath effects. Therefore, data collected with this bandwidth are used for further analysis.

\begin{figure}[t]
    \centering
    \includegraphics[width=0.9\linewidth]{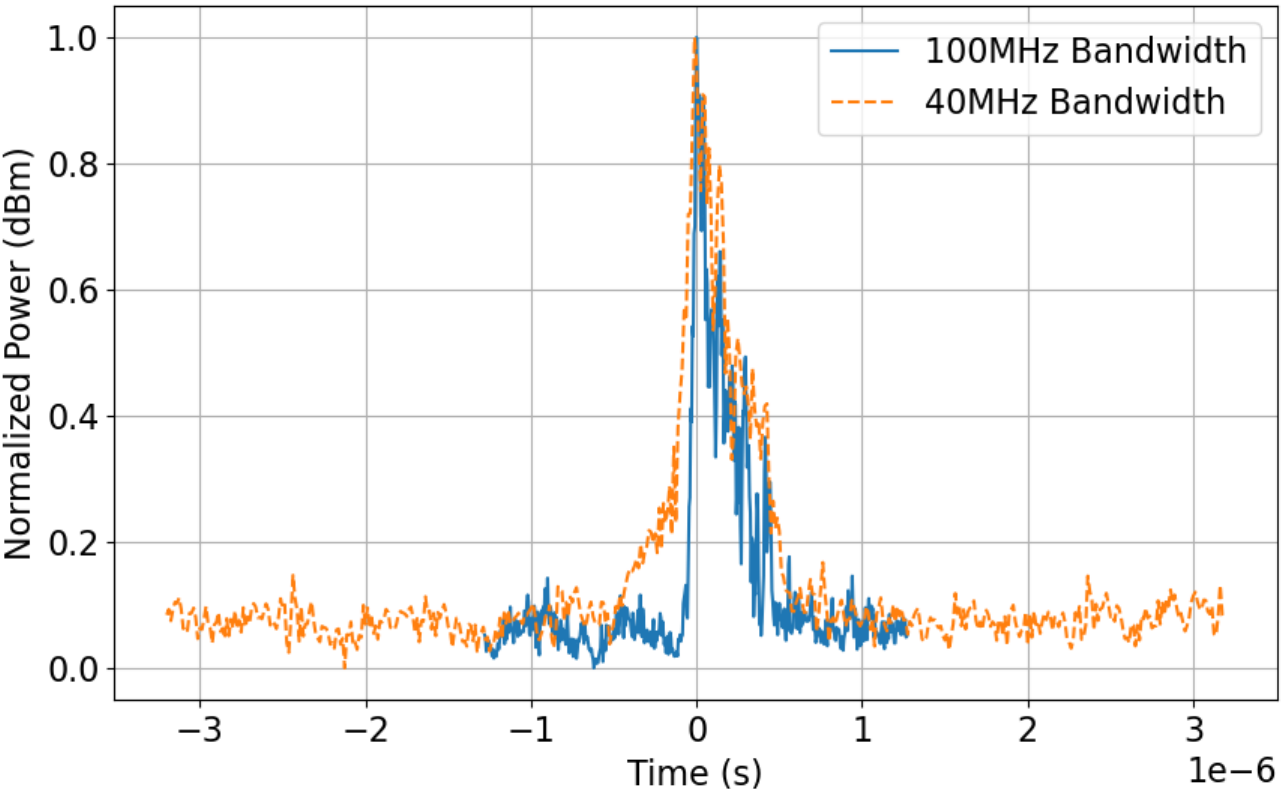}
    \caption{Normalized CIR collected by Keysight under different bandwidths.}
    \label{fig:keysight-cir}
\end{figure}

\subsection{Example of Synthetic CIR}
\label{app: wi-cir}
Fig.~\ref{fig:wi-cir} shows one CIR example generated from the WI platform, where the horizontal axis is the time of arrival and the vertical axis is the corresponding amplitude, providing a strong data foundation for further validation.

\begin{figure}[t]
    \centering
    \includegraphics[width=0.9\linewidth]{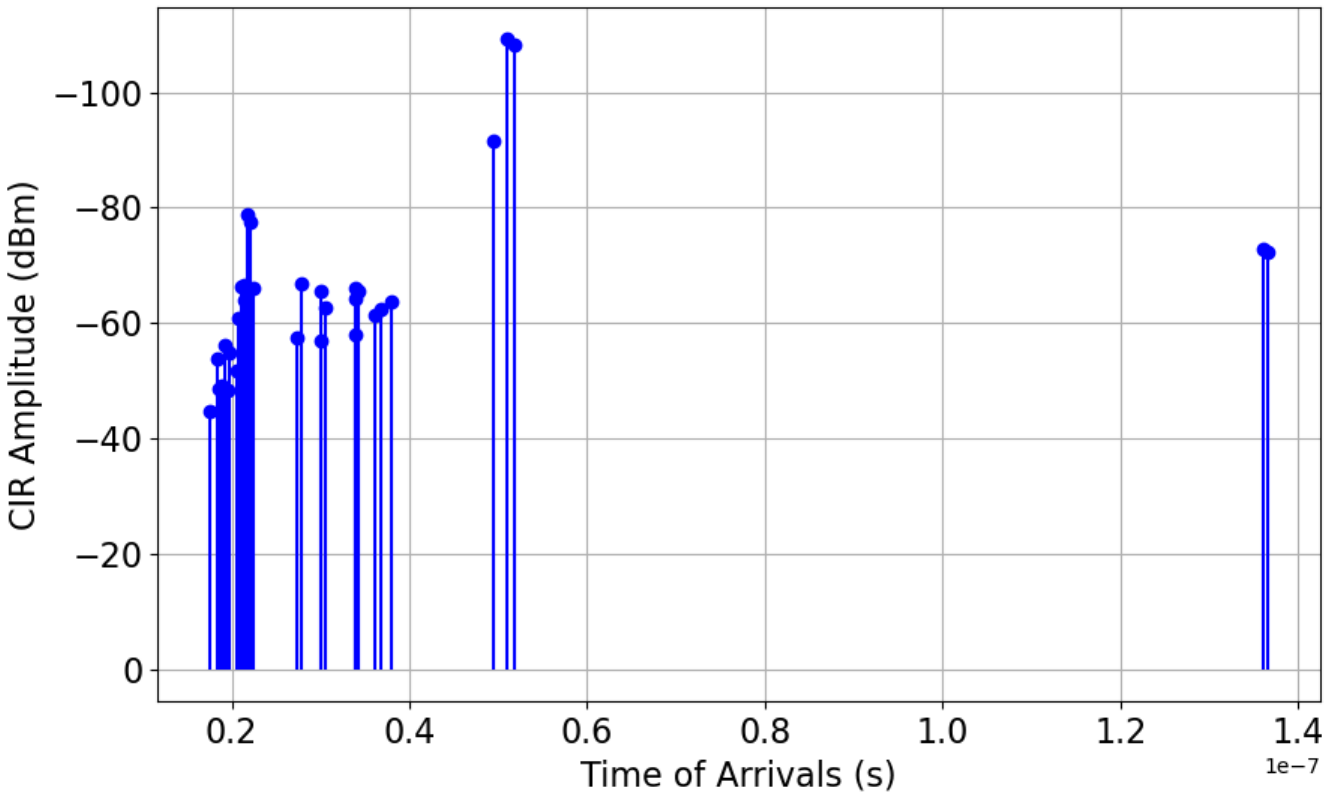}
    \caption{CIR simulated by WI.}
    \label{fig:wi-cir}
\end{figure}

\subsection{Model Configurations}
\label{sec: model_configurations}

We summarize the detailed configurations of all models used in the experiments as follows.
\begin{itemize}
    \item KNN~\cite{bahl2000radar}: We employ the Euclidean distance metric to identify the $K=5$ nearest reference points in the signal space corresponding to the estimated TP. The estimated location is then computed as the average of the selected reference points.
    
    \item WKNN~\cite{WKNN2020indoor}: Similar to KNN, we utilize the Euclidean distance metric to select the $K=5$ closest reference points. Additionally, we calculate the cosine similarity between the features of each fingerprint, with the weights for the reference points determined by these similarity scores.

    \item CNN~\cite{ghozali2019indoor}:  This model consists of three convolution-pooling layers, each containing a convolution layer with $16@3\times 1$ kernels and a pooling layer with $16\times1\times1$ channel, followed by two fully connected hidden layers, each with 200 neurons. The CNN is trained for a maximum of 10000 epochs using a full batch size, with the Adam optimizer and a learning rate of 0.001.

    \item MLP~\cite{gao2022metaloc}: This model comprises four hidden layers, each containing 200 neurons. The MLP is trained for a maximum of 5000 epochs using a full batch size, with the Adam optimizer and a learning rate of 0.01.

    \item MetaLoc~\cite{gao2022metaloc}: The neural network architecture for MetaLoc is identical to that of the MLP. The inner loop's step size is set to $\alpha = 0.0001$, while the outer loop's step size is $\beta = 0.001$. Meta-training is conducted over 10000 epochs, followed by a meta-testing phase lasting 100 epochs.

    \item AGNN (ours): For the DLM, we set $D_{M_1}=D_{M_2}=20$. In the ALM, we configure $F_A=50$. In the two-layer MAGL architecture, $F_{att}=50$ and $D_k=100$ are utilized for each layer. The AGNN model is trained for a maximum of 5000 epochs with a full batch size, using the Adam optimizer with a learning rate of 0.01.

    \item AGML (ours): The neural network configuration for AGML is consistent with that of AGNN. The inner loop's step size is set to $\alpha = 0.0001$, while the outer loop's step size is $\beta = 0.001$. The model undergoes meta-training for 10,000 epochs, followed by meta-testing for 100 epochs.
\end{itemize}

\subsection{The Approximated Step Function}
\label{app: appro_step_func}
As shown in Fig.~\ref{fig: appro_step_func}, the approximated step function, employing ReLU and tanh, yields a notably sharper transition on one side and effectively truncates values to zero on the other side, consequently achieving a more accurate approximation of the step function than the sigmoid function. Moreover, its derivative can be obtained everywhere (In the standard definition of ReLU, the derivative at 0 is typically taken to be 0.), which makes it feasible to compute gradients with respect to the trainable threshold.

\begin{figure}[t]
	\centering
	\includegraphics[width=0.9\linewidth]{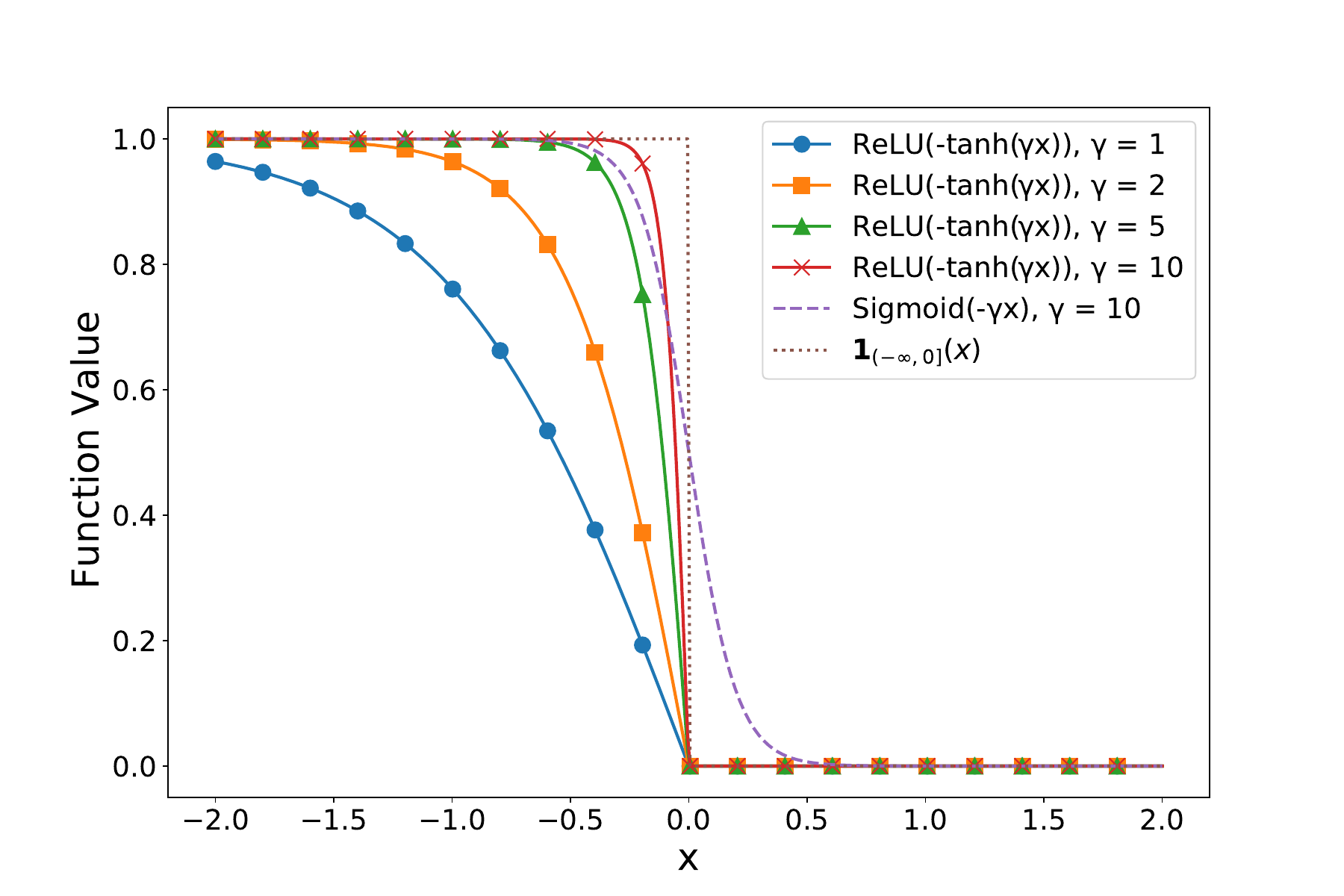}
	\caption{Comparison between sigmoid function, step function, and approximated step functions.}
	\label{fig: appro_step_func}
\end{figure}

\fi

\end{document}